\documentclass[10pt,journal,compsoc]{IEEEtran}
\ifCLASSOPTIONcompsoc
  \usepackage[nocompress]{cite}
\else
  \usepackage{cite}
\fi
 
\usepackage{mathtools,amssymb} 
\usepackage{xcolor}
\usepackage{adjustbox}
\usepackage{array}
\usepackage{float}
\usepackage{booktabs}
\usepackage{textcomp}
\usepackage{graphicx}
\usepackage{algorithm}
\usepackage{algorithmic}
\usepackage{multirow}
\usepackage{subfigure}

\newcommand{\pan}[1]{\textcolor{black}{#1}}

\usepackage[pagebackref=true,breaklinks=true,letterpaper=true,colorlinks,bookmarks=false]{hyperref}
\ifCLASSINFOpdf
\else
\fi
\begin{document}

\title{A Framework of Meta Functional Learning  for Regularising Knowledge Transfer}

\author{Pan~Li,\ %~\IEEEmembership{Member,~IEEE,}
        Yanwei~Fu\ %~\IEEEmembership{Fellow,~OSA,}
        and~Shaogang~Gong%~\IEEEmembership{Life~Fellow,~IEEE}% <-this % stops a space

\IEEEcompsocitemizethanks{\IEEEcompsocthanksitem Pan Li and Shaogang Gong are with the school of Electrical Engineering and Computer Science, Queen Mary University of London, London,
UK, E1 4NS.\protect %\\
E-mail: \{pan.li, s.gong\}@qmul.ac.uk
\IEEEcompsocthanksitem Yanwei Fu is with the School of Data Science,
Fudan University, and Shanghai Key Lab of Intelligent Information
Processing, Fudan University. Yanwei Fu is also with the MOE Frontiers Center for Brain Science, Fudan University. \protect %\\
E-mail: yanwei.fu@fudan.edu.cn
}% <-this % stops an unwanted space
}
\IEEEtitleabstractindextext{%
\begin{abstract}
Machine learning classifiers' capability is largely dependent on the scale of available training data and limited by the model overfitting in data-scarce learning tasks. 
To address this problem, this work proposes a novel framework of Meta Functional Learning (MFL) by meta-learning a generalisable functional model from data-rich tasks whilst simultaneously regularising knowledge transfer to data-scarce tasks. 
The MFL computes meta-knowledge on functional regularisation generalisable to different learning tasks by which functional training on limited labelled data promotes more
discriminative functions to be learned. 
Based on this framework, we formulate three variants of MFL: MFL with Prototypes (MFL-P) which learns a functional by auxiliary prototypes, Composite MFL (ComMFL) that transfers knowledge from both functional space and representational space, and MFL with Iterative Updates (MFL-IU) which improves knowledge transfer regularisation from MFL by progressively learning the functional regularisation in knowledge transfer. 
Moreover, we generalise these variants for knowledge transfer regularisation from binary classifiers to multi-class classifiers.
Extensive experiments on two few-shot learning scenarios, Few-Shot Learning (FSL) and Cross-Domain Few-Shot Learning (CD-FSL), show that meta functional learning for knowledge transfer regularisation can improve FSL classifiers.
\end{abstract}

}

\maketitle

\IEEEdisplaynontitleabstractindextext

% \fi
\IEEEpeerreviewmaketitle

\IEEEraisesectionheading{\section{Introduction}\label{sec:introduction}}

\IEEEPARstart{T}{he} success of current deep architectures benefits a great deal on
representation learning, in the sense of learning ``big models''
of richer representations for many tasks. Recent developments on self-supervised
learning, or models trained on very large-scale data~\cite{devlin2018bert,brown2020language},
seem to suggest that powerful and universal representations could
be learned for all tasks in all domains.

Given a universal feature extractor, can a good classifier for a
particular task be effectively learned from only a few labelled examples
of that task? Having a good universal representation does not guarantee
fitting generalisable hypotheses of different individual tasks from
a few labelled samples. For a Few-Shot Learning (FSL) task, many researchers
had devoted their efforts in addressing the severe overfitting problem
resulting in inferior classification accuracy and generalisation on novel
categories~\cite{ravi2016optimization,ye2020few}. Typical FSL settings
\cite{chen2019closer,gidaris2018dynamic,dhillon2019baseline} assume that given a large amount of labelled
data on source/base tasks, and few labelled data on target/novel tasks,
a FSL algorithm can learn good hypotheses on novel tasks. Moreover,
one may further consider Cross-Domain Few-Shot Learning (CD-FSL) when
the source and target tasks are from significantly different semantic
domains~\cite{Tseng2020Cross-Domain,guo2020broader}.

\begin{figure}[t]
\centering{}
\includegraphics[width=0.48\textwidth]{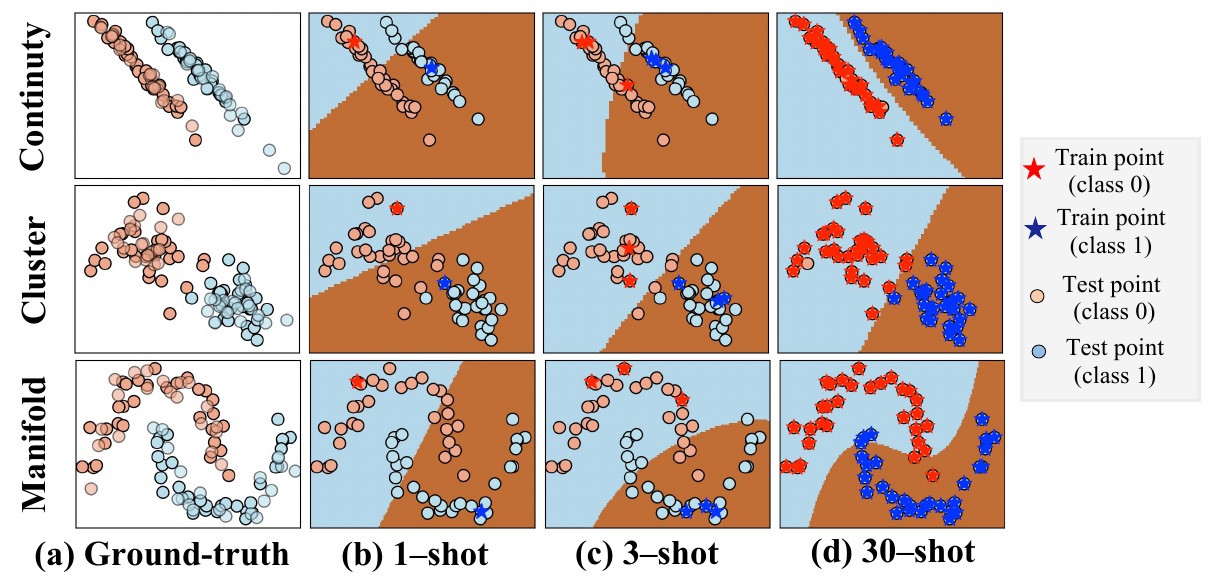}
\vspace{-0.05in}
\caption{The illustration of hypotheses learned with $k$-shot data on binary classification tasks with continuity, cluster and manifold distributions.
Plots (a) are the ground-truth data distributions, and (b-d) represent
the hypotheses learned with 1/3/30-shot data. 
Without efficient data training, the hypotheses (plots (b)) fail to learn the ground-truth data distributions, whilst the hypotheses (plots (c-d)) are progressively capable to learn them with the increased regularisation knowledge deriving from the labelled data.
Best viewed in color. \label{fig:An-illustration-of}}
\end{figure}

Given a learned representation from richly labelled data, we consider
that the underlying data distribution should follow the \emph{continuity},
\emph{cluster}, and \emph{manifold} assumptions, as in Semi-Supervised
Learning (SSL)~\cite{chapelle2009semi}. Figure~\ref{fig:An-illustration-of}
illustrates this phenomenon from both SSL and supervised learning.
Hypotheses learned from larger amount of examples (richer) are favoured
than those trained by fewer examples. Moreover, good hypotheses should
prefer geometrically simpler decision-boundaries and encourage points
in the same cluster to have the same label. This should be a
general principle for task-agnostic patterns of a hypothesis.

% yw: meta functional learning 
In a hypothesis/functional space, we aim to learn gradually task-agnostic
patterns of change in fitting hypotheses to training data from few
to many labelled examples. In particular, the latent knowledge of
task-agnostic patterns of change in a hypothesis fitting process is
to be learned as a {\em functional}, estimated from a family of
richly labelled data on source tasks that simultaneously satisfies
new hypotheses of the same/similar family of functional generalisable
to learning new target tasks. To that end, we introduce a framework with meta-learning strategy to learn this functional, called {\em Meta Functional Learning}
(MFL).

% yw: functional regularization 
Essentially, our MFL framework learns a functional regularisation on how to
best fit new hypotheses on scarcely labelled novel tasks according
to how to best fit hypotheses on richly labelled base tasks,
thus imposing penalties (constraints) on excessive optimisation (overfit)
in fitting the novel hypotheses. 
Particularly, given the task of learning a novel hypothesis from scarcely labelled data, our functional encourages {\em a process} of learning the hypothesis by approximating the learning process of richly labelled data, from which it favours to satisfy the underlying data distribution principles of continuity, cluster, and manifold.
The functional from MFL captures model learning {\em \textbf{regularisation knowledge}} from source data and transfers
it to guide the FSL of novel tasks.
Our approach to knowledge transfer as learning regularisation ({\em how to learn}) differs fundamentally to other existing methods of knowledge transfer on {\em what to learn}, e.g. representations in FSL.
Figure~\ref{fig:An-illustration-of} illustrates our idea of MFL that learns a task-agnostic, transferable and generalisable functional, a function in the functional space, to remit the overfitting problem in hypothesis optimisation given scarcely labelled data.   

We explore a meta-learning paradigm to learn a
functional of meta-knowledge as the regularisation of learning process.
Essentially, MFL first samples many functional episodes to craft a functional
set of paired classifiers trained by the corresponding few and many labelled data, respectively. MFL is learned to predict the functional of many labelled data, given the input functional from few labelled data. It achieves the meta-knowledge learning/transfer through functional regularisation. This is the vanilla MFL proposed in our conference version~\cite{Panijcai2021}. 

Based on the understanding of vanilla MFL~\cite{Panijcai2021}, we generalise it and improve the formulations by introducing new variants:
1) We explore the information from prototypes (MFL-P) as the classes' examples can provide the relative positional relationship of classes to help the functional learning in functional space;
2) To learn more regularisation knowledge from multiple source information, we additionally introduce a functional in representational space and naturally combine it with the functional in functional space, formulating a Composite MFL (ComMFL);
3) As the functional learned from one block of MFL contains limited capacity for transferring the regularisation knowledge,  we employ an iterative update strategy to connect a sequence of basic module blocks (MFL-IU) to progressively learn the generalisable regularisation knowledge.

Moreover, we consider a more challenging learning problem, that is, the functional learning for a multi-class classifier which has higher dimensions of parameters, larger functional space and extra inter-class relationships compared with that for a binary classifier.
As a trail, we generalise our MFL methods to multi-class classifiers by introducing an outer loop for MFL to capture a functional from more episodes.

We summarise our contributions as follows.

\begin{itemize}
\item We formulate knowledge transfer in few-shot learning as a problem
of transfer learning regularisation (how to learn)
rather than knowledge transfer in representation (what to learn).
This problem is solved by 
\pan{a meta functional learning (MFL) framework. 
\item We introduce three variants of MFL, i.e. a MFL-P that learning a functional with auxiliary information from prototypes, a ComMFL which learns a composite functional with wider regularisation knowledge from functional space and representational space, and a MFL-IU employing an iterative update strategy for MFL that aims to gradually improve the  classifier's learning ability through
the transfer of functional regularisation. 
\item We generalise our MFL methods to the functional learning from binary classifiers to multi-class classifiers, and introduce a readily ensemble method to improve the robustness of classifiers.
}
\item We apply the MFL to both the standard few-shot learning and the cross-domain
few-shot learning problems. We provide comprehensive experiments on
\emph{mini}ImageNet, CIFAR-FS, CUB, Cars and Places to validate the effectiveness of MFL
and its variants in improving FSL by minimising model overfit.
\end{itemize}

\section{Related Work}
% \subsection{Functional learning \todo{need to change: functional gradient optimisation}}
\subsection{Functional Optimisation}
Functional optimisation can be regarded as learning to optimise the function.
Many works put efforts on learning functional gradients to optimise neural networks by functional gradient optimisation~\cite{johnson2019framework,garg2020functional,titsias2020functional,johnson2020guided} or functional gradient boosting~\cite{khot2011learning_boosting,nitanda2018functional_boosting,nitanda2020functional_boosting}.
% The functions are always the basic function for knowledge learning from data, e.g. the representation function~\cite{garg2020functional} from data and the computed gradient~\cite{johnson2020guided,johnson2019framework} for data.
%
For example, \cite{johnson2020guided} computes a guide function to optimise the gradient function, formulating a functional gradient optimisation method.
Apart from optimisation for gradient, Garg et. al.~\cite{garg2020functional} presented a functional optimisation on representation and unifies several self-supervised approaches as a framework to impose a regularisation on the representation via a learnable function using unlabeled data. 
Rather than learning a functional to optimise the gradient or representation, we aim to meta-learn a functional to regularise the knowledge transfer for classifiers.

\subsection{Regularisation}
Regularisation is an important technique to improve the generalisation ability of machine learning models in both traditional classification methods~\cite{lee2006efficient_reg_lr,ng2004feature_reg,cawley2007sparse_reg_lr,cawley2007preventing_reg_svm} and currently popular deep learning methods~\cite{lang1990dimensionality_weight_decay,gal2016dropout,gouk2021regularisation_net}.
Lee et al.~\cite{lee2006efficient_reg_lr} and Andrew Y. Ng~\cite{ng2004feature_reg} have investigated the effects of $L_1$ and $L_2$ regularisation for improving the generalisation ability of Logistic Regression (LR) and Support Vector Machine (SVM).
Moreover, they give some theoretical proof that regularisation can reduce the generalisation error bound of classifiers.
For deep learning methods, some classical regularisation techniques have been widely used, such as weight decay~\cite{lang1990dimensionality_weight_decay} and dropout~\cite{gal2016dropout}.
MetaReg~\cite{balaji2018metareg} proposed to explicitly meta-learn a regularisation function for domain generalization. 
Related to these regularisation techniques, our work is more focused on improving the generalisation ability of traditional classifiers, e.g. LR and SVM, by exploring a learnable and implicit regularisation module equipped with deep learning method.
% mainly refer to~\cite{garg2020functional,titsias2020functional}

\subsection{Meta Learning}
Recently, the idea of \textit{meta-learning} or \textit{learning to learn}\cite{thrun2012learning_to_learn} has been exploited by the machine learning community, as it shows a promise to achieve close to human-level recognition generalisation potential in a controlled setting~\cite{wu2018meta,nips2018MetaGAN,li2017meta-sgd}.
In~\cite{balaji2018metareg}, the authors used meta-learning to train a regularisation item for neural network optimisation across domains and demonstrated the benefits to addressing the domain generalization problem.
More works~\cite{finn2017model,wu2018meta,li2017meta-sgd} are related to few-shot learning. MAML~\cite{finn2017model} is one of the representatives dealing with few-shot learning task by learning to learn a generalisable initialisation parameters for networks. 
Different from  MAML that only takes meta-learning an initialisation, Meta-SGD~\cite{li2017meta-sgd} presented a method with much higher capacity by additionally learning the meta-learner updating direction and learning rate for few-shot tasks. 
Rather than learning the optimisation process with meta learner, MeLA~\cite{wu2018meta} is a simpler meta-learner to directly generate model parameters for few-shot tasks.
In our work, we also use meta-learning to solve the learning problem with limited labels but aim to train a meta-learner for transferring the implicit regularisation knowledge for few-shot tasks.
% Recently, there has been a lot of interest in applying such strategies for deep neural networks.

\subsection{Transfer Learning}
Transfer learning aims to leverage the prior knowledge from source training data to address the target tasks where only limited labelled data are available~\cite{pan2010survey,torrey2010transfer}. 
A typical method~\cite{mahajan2018exploring,cui2018large} for transfer
learning is fine-tuning a model pre-trained on a well-labelled base
dataset with limited novel target data.
Another approach tries to reduce the distance between the
distributions of a source domain and the target domain so to better
transfer the knowledge learned from the source
domain~\cite{liang2020we}\cite{li2020maximum}.
These transfer methods are widely used for domain adaptation, which
assumes that the source and target domain share the same label space. 
In practice, most source and target domains do not share the same
label space, giving rise to the learning problems of open set
recognition~\cite{scheirer2012openset} and few-shot
learning~\cite{qi2018low}. 
Our work aims to solve some transfer learning problems with limited labelled data, e.g. few-shot learning, cross-domain few-shot learning.
% supervised domain adaptation, 

\subsection{Few-Shot Learning}
Few-shot learning is a task requiring fast recognising novel classes with very limited corresponding labelled samples.
Existing FSL methods can be broadly characterized as follows.
 1) Metric-based methods learn a common feature
space where categories can be distinguished with each other based on a
distance metric, and then infer labels for query data with a nearest
neighbor classifier~\cite{snell2017prototypical} or a separate learnable
similarity metric~\cite{sung2018learning}. 2) Gradient-based methods
design the meta-learner as an optimiser that is learned to update model
parameters. These approaches aim to learn good initialised parameters
for a network so that the classifiers for novel classes can be learned
with several gradient update steps on few labelled examples~\cite{finn2017model,ravi2016optimization,li2021semi}.
3) Weight generation methods learn to generate classification weights
for novel classes. A typical generation method directly
predicts the classification
weights from the activation statistics of their categories~\cite{gidaris2018dynamic,qi2018low}. Besides, some work
try to generate better classification weights with denoising auto-encoders
for weights reconstruction~\cite{gidaris2019generating} or looking
into the mutual information between generated weights and support/query
data~\cite{guo2020attentive}. Different from existing work to generate
weights from the activations of a feature extractor, we aim to investigate
the function learning update dynamics (a functional) which is not
limited to backbone training strategies.

\subsection{Model Transformation and Composition}
Our investigation on knowledge transfer by functional regularisation is related to previous works on model transformation and composition, in particular, a model regression network with MLP architecture for learning a generic, category agnostic transformation from small-sample models to the underlying
large-sample models~\cite{wang2016learning}. 
Subsequently, a MetaModelNet~\cite{wang2017learning} was proposed for transferring the model dynamic from head classes to tail classes in long-tail recognition problem. 
Functional gradient learning~\cite{johnson2019framework} was explored to learn the composition of functions and an incremental strategy was adopted for gradually learning a generator network.
Our work is partly inspired by these works but we expand the existing works to a new method of meta functional learning to construct generalisable learning regularisation knowledge capable of guiding `infant' functions to become `mature' functions in a process of function update.

\section{Meta Functional Regularisation}
\subsection{Problem Definition.}
Thoughout the paper, we use $\mathbf{I}$ to denote the image data, $y$ represents the corresponding label. 
we learn a representation function $\psi:\mathbf{I}\rightarrow x$ and $x\in\mathcal{R}^p$,
and a classifier $f:\psi\left(\mathbf{I}\right)\rightarrow y$. 
And the corresponding representational space and functional space are represented as $\mathcal{H}_\psi$ and $\mathcal{H}_f$.

In the transfer learning scenario, we consider a large-scale labelled source/base image-label pair set
$D_{src}=\left\{ \mathbf{I}_{j,}y_{j}\right\} _{j=1}^{M}$, $y_{j}\in\mathcal{C}_{base}$,
and a small labelled novel/target image set $D_{nov}=\left\{ \mathbf{I}_{j,}y_{j}\right\} _{j=1}^{N}$,
$y_{j} \in \mathcal{C}_{nov}$, from a base $\mathcal{C}_{base}$ and
a novel category $\mathcal{C}_{nov}$ respectively. On $D_{src}$,
we learn a representation function $\psi$,
and then we learn a classifier $f_{\phi}$,
where $\phi$ is the parameter of $f$. A common practice in deep
learning is  end-to-end optimising $\psi$ and $f$ by formulating
a multi-class classification problem over $D_{src}$ with a cross-entropy
loss. We utilise this process here to compute a feature representation
$\psi$.

\noindent \noindent \textbf{Functional learning.}
Our goal is to learn to fit a \textit{functional regularisation,}
$\mathcal{T}:f_{\phi}\rightarrow f_{\tilde{\phi}}$.
Specially, the input $f_{\phi}\left(\psi\left(\mathbf{I}\right)\right)$
is a classifier fitted by few labelled samples, and $\mathcal{T}\left(f_{\phi}\right)$
aims at approximating the corresponding function $f_{\tilde{\phi}}$ with regularisation knowledge learned from many labelled examples. 
We use $\phi$ and $\tilde{\phi}$ to denote the parameters learned by few and many labelled examples.

\subsection{Insights of Functional Regularisation}
\noindent \noindent \textbf{Model Dynamics, and Functional Regularisation.}
From the learning principles of risk minimization~\cite{vapnik1992principles_erm} and given a binary classification task with dataset $\mathcal{D}_S=\{x_i,y_i\},i=1,...,n$, we can obtain $m$ different subsets $\{S_1, ...S_k..., S_m\}$ according to the number $k$ of training instances.
For every subset $S_k$, we can use the same classification algorithm to train a set of classification models/functions $\{f_1^*,...,f_k^*, ..., f_m^* \}$ by Empirical Risk Minimization (ERM) with Eq.~(\ref{eq:erm}).
\begin{equation}
\label{eq:erm}
R_{erm}(f) = \min_{f\in \mathcal{H}_f}\ \frac{1}{n} \sum_{i=1}^{n} L_c(y_i,f(x_i)),
\end{equation}
where $L_c$ is a loss function to compute the errors.
This model can also be optimised by Structural Risk Minimization (SRM) in Eq.~(\ref{eq:srm}) which uses a regularisation term $J(f)$ to increase the model's generalisation ability and a coefficient $\lambda$ to balance the learning of $R_{erm}(f)$ and $J(f)$.
\begin{equation}
\label{eq:srm}
R_{srm}(f) = \min_{f\in \mathcal{H}_f}\ \frac{1}{N} \sum_{i=1}^{N} L_c(y_i,f(x_i)) + \lambda J(f).
\end{equation}
Suppose that $m$ is large enough, we can yield an infinite set of optimal functions, which can form a functional space $\mathcal{H}_f$. In this functional space, vector $f_0^*$ represents the function without any data training. Thus it can be viewed as a randomly initialised vector in the functional space. With the increase of $k$, function $f_k^*$ can be viewed as the model dynamics in the functional space towards the optimal function $f_m^*$. To this end, we simply formulate the model convergence of model dynamics in the functional space.

\begin{figure}[t]
\centering
\includegraphics[scale=0.56]{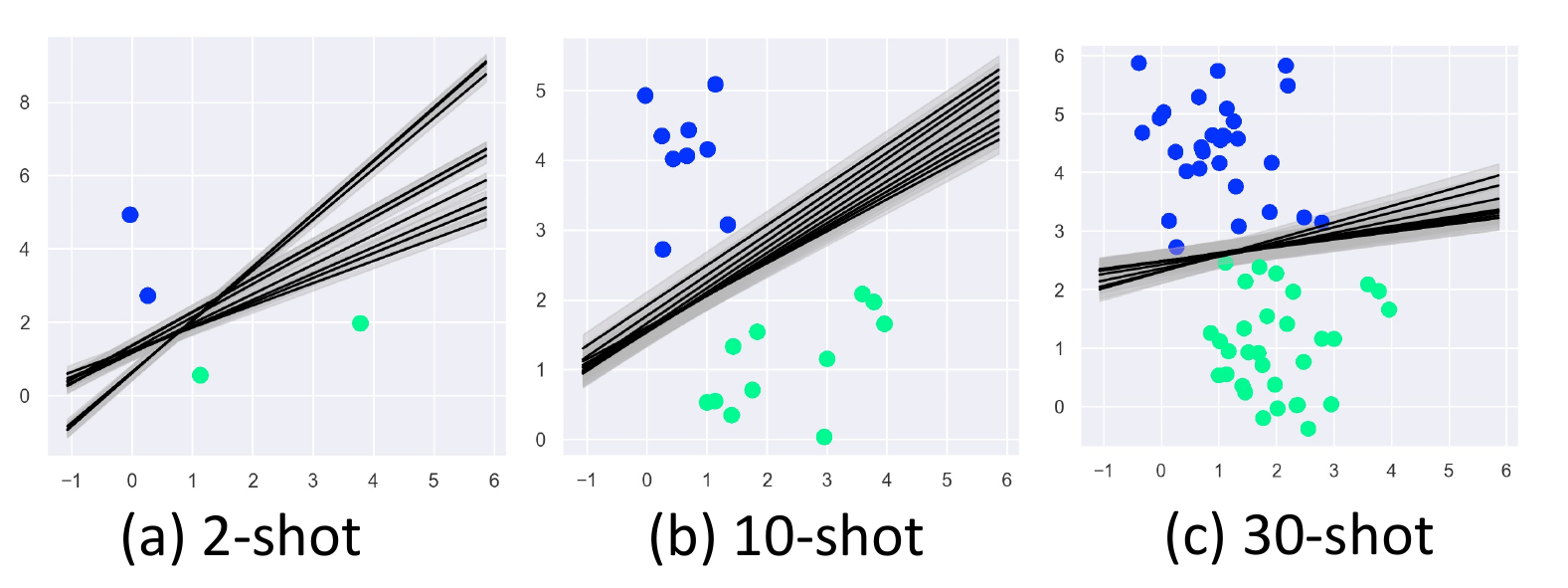}
\vspace{-0.2in}
\caption{An illustration of the gradual constricting process in functional space with the increase of shots. Plots (a, b and c) show the classifier boundaries using the Logistic Regression with different hyper-parameters trained on 2, 10 and 30 labelled samples. The blue points and green points present the training samples from two classes, and the black lines are the trained classifier boundaries with the training instances.} \label{fig:hypotheses}
\end{figure}

\textit{So what is the `implicit' knowledge of learned  by model dynamics in the functional space?} Here, we try to intuitively explain it from the perspective of knowledge regularisation. Suppose that we just have two data points in a representational. And there are many hypotheses that are able to well separate these two points.
As in Fig.~\ref{fig:hypotheses}, with the increase of training samples, the feasible space of classification functions will become more and more narrow. The extreme case is that the training samples are infinite and can well show the data distribution, the feasible space for classification functions will be narrowed in a small space. The changing of the feasible space with the increase of training instances can be viewed as the pruning or constricting process in the functional space.

\subsection{Insights of Meta Functional Learning}
% \noindent \textbf{Meta learning task-agnostic functional. } % Estimation 
Rather than directly regressing $\mathcal{T}$ by a crafted functional
set, we adopt a meta-learning strategy here. In principle, such a
strategy helps cover a distribution of related tasks, sampled by episodes,
and thus mimicking the predicting future functions from different
domains. Our insight is that: {\em despite the data may be different
between the source and target domains, the underlying 
patterns of change in fitting hypotheses to training data from few to many labelled examples, should be
in principle, the same, or similar at least.} % although with different values}. 
The functional $\mathcal{T}$ learned to represent such
meta-knowledge of model convergence in one domain, could be generalisably
applied to a novel domain. To that end, $\mathcal{T}$ should
be learned in a task-agnostic manner. 

\noindent \textbf{Learning task-agnostic knowledge transfer.} Our
empirical study (in Fig.~\ref{fig:An-illustration-of}) shows the task-agnostic knowledge, i.e. the meta-knowledge of \emph{functional regularisation}, extracted from a family of source tasks,
could potentially be utilised to improve the generalisation of new
tasks from that family. Particularly, to learn a regularisation knowledge
transfer, we adopt the meta-learning strategy to learn the functional
$\mathcal{T}$ over multiple learning episodes of the source tasks,
sampled from base categories $\mathcal{C}_{base}$. Then the learned
functional $\mathcal{T}$ is \textcolor{black}{generalised
and applied to tasks in target dataset $D_{nov}$.} 

\begin{figure*}[t]
\centering{}
\includegraphics[scale=0.38]{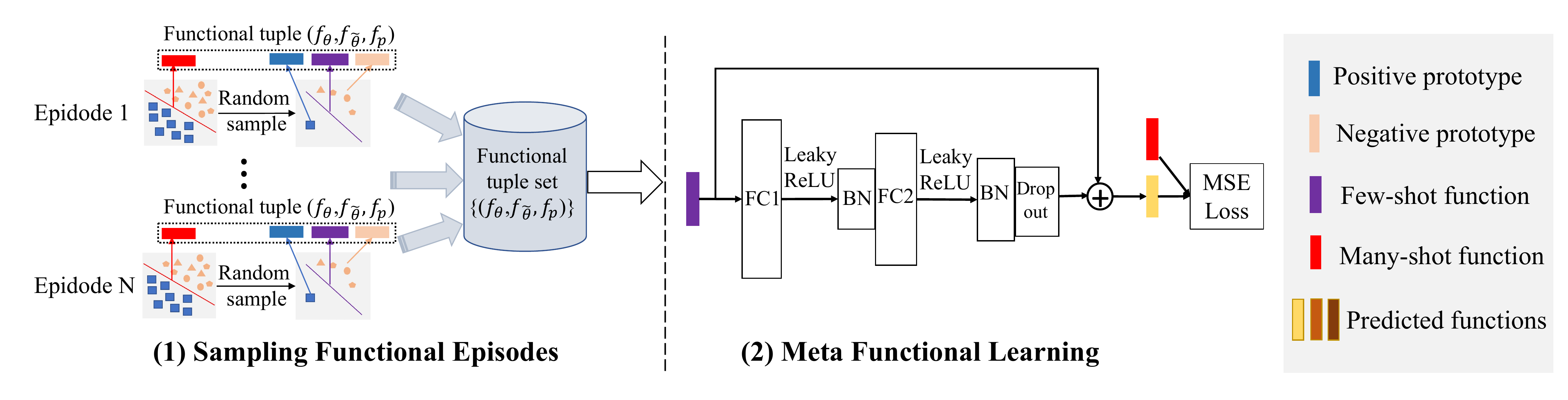}
\vspace{-0.15in}
\caption{The overall model design for Meta Functional Learning (MFL). (1) The functional episodes are sampled from the base dataset, and compute the functional tuple, \emph{i.e.} the prototypes $f_{p}$, the few-shot classifier/function $f_{p}$ and the many-shot classifier/function $f_{\hat \phi}$; (2) The meta functional learning is used for predicting a many-shot function given few-shot function and prototypes.
\label{fig:The-pipeline-of}}
\end{figure*}

\section{Meta Functional Learning} 
In this section, we first introduce a general framework, then develop and analyse the corresponding algorithms.

\subsection{Methodology in a Nutshell}
As our approach to MFL focuses on learning the functional $\mathcal{T}$ for classifiers on a fixed representational space, we first train a representor to extract the representations from images.
Specifically, we follow the traditional mini-batch training strategy in~\cite{wang2020instance} and use the cross-entropy loss to pre-train a representor on source dataset $D_{src}$.
After training a representor, the functional $\mathcal{T}$ can be learned with the following two steps:

\noindent \textbf {Step 1: Functional Initialization (Sec.~\ref{subsec:sampling}).}
For a classifier $f_{\phi}$ trained on task $T$ containing limited data, the function is not the ideal one due to the over-fitting problem. 
While the ideal function is hard to compute since the true distribution of task $T$ is not available. Here we approximately compute the ideal function $f_{\hat \phi}$ by training classifier with more available data of the classes in task $T$ .

\noindent  \textbf {Step 2: Functional Learning (Sec.~\ref{subsec:scratch}).}
The functional $\mathcal{T}$ helps the function $f_{\phi}$ computed with limited label to approach the ideal function.
Therefore, one intuitive way is to learn a $\mathcal{T}$ to capture this knowledge guiding the function $f_{\phi}$ to the approximate ideal function $f_{\hat \phi}$.

To learn a task-agnostic and generalisable functional $\mathcal{T}$ by using a meta-learning paradigm, a common practice is to optimise the functional by iteratively computing step 1 and 2 process. Unfortunately, such an exhaustive and iterative updating process demands frequently initialising functionals, and thus is difficulty for parallel-computing in batches.
Therefore, we design a simpler Meta Functional Learning (MFL) framework to train the functional. This alternative pipeline of MFL can be illustrated as Fig.~\ref{fig:The-pipeline-of}:
(1) We sample the functional episodes to train $\mathcal{T}$ (Sec.~\ref{subsec:sampling}); and
(2) we learn $\mathcal{T}$ by different strategies (Sec.~\ref{subsec:scratch}).

\subsubsection{Sampling Functional Episodes \label{subsec:sampling}}
\label{sec:vanilla-sampling}
Given the trained representator $\psi$, the goal of this step is
to craft the paired functional set $\mathcal{F}_{\mathcal{T}}=\left\{ \mathcal{F}_{\mathcal{T}}^{\left(b\right)}\right\} $
on $D_{src}$ and the class $b\in\mathcal{C}_{base}$; and we denote
$\mathcal{F}_{\mathcal{T}}^{\left(b\right)}=\left\{ \left(f_{\phi}^{\left(b\right)},f_{\tilde{\phi}}^{\left(b\right)},f_{p}^{\left(b\right)}\right)\right\} $\textcolor{black}{,
where $f_{\phi}^{\left(b\right)}$ and $f_{\tilde{\phi}}^{\left(b\right)}$
are the classifiers of class $b$, trained by few and many examples,
respectively; 
% camera ready
and $f_{p}^{\left(b\right)}$ represent the prototypes of the positive class $b$
and other negative classes, 
computed by few labelled examples
which are used for training $f_{\phi}^{\left(b\right)}$ .  }

\textcolor{black}{The sampled functional episodes include different
classes in $\mathcal{C}_{base}$. This will help our meta functional
learning algorithm to learn task-agnostic functional $\mathcal{T}$.
Specifically,  for class $b$ ($b\in\mathcal{C}_{base}$), we compute functional tuple set $\mathcal{F}_{\mathcal{T}}^{\left(b\right)}=\left\{ \left(f_{\phi}^{\left(b\right)},f_{\tilde{\phi}}^{\left(b\right)},f_{p}^{\left(b\right)}\right)\right\} $.
For each tuple, $f_{\tilde{\phi}}^{\left(b\right)}$ is trained by
the set of positive examples $\left\{ \psi\left(\mathbf{I}_{j}\right),y_{j}=b\right\} $,
} i.e. all images in class
$b$, and negative examples $\left\{ \psi\left(\mathbf{I}_{j}\right),y_{j}\neq b\right\} $
by randomly sampling from other classes. To obtain the set of tuples,
this process is randomly repeated for $M_{l}$ times. To compute $f_{\phi}^{\left(b\right)}$,
we take $s$ samples and $k\text{\ensuremath{\times}}s$ samples
from class $b$ and other classes. For each $f_{\tilde{\phi}}^{\left(b\right)}$,
we randomly sample samples $M_{f}$ times and use different hyper-parameters
to train the classifiers $f_{\phi}^{(b)}$ for increasing their diversity.

\subsubsection{Learning from the Scratch \label{subsec:scratch}}
\label{sec:vanilla-mfl}
As a vanilla instantiation of our MFL framework, we adopt the $f_{\phi}^{\left(b\right)}$
by a vanilla binary classifier for class $b$, and the generalised
multi-class scenario (one \emph{ vs.} all setting).
% is also extended in the experiments.
%
We utilise the Logistic Regression (LR) classifiers
here, and\textcolor{black}{{} $f_{\phi}^{\left(b\right)}$ and $f_{\tilde{\phi}}^{\left(b\right)}$}
are the corresponding vectors of LR parameters. For the vanilla MFL, we
directly learn $\mathcal{T}:f_{\phi}\rightarrow f_{\tilde{\phi}}$ in functional space.

Given the functional sets $\mathcal{F}_{\mathcal{T}}$, we design
a meta functional learning mechanism to learn the functional regularisation
$\mathcal{T}$. For any given class $b$, the objective of our MFL
is to approximate the ground-truth output $f_{\tilde{\phi}}^{\left(b\right)}=\mathcal{T}\left(f_{\phi}^{\left(b\right)}\right)$.
We introduce Mean Square Error (MSE) to measure the difference of
parameter vectors $\left(f_{\phi},f_{\tilde{\phi}}\right)$
as, 
\begin{equation}
{
l_{\tau}=\mathbf{E}_{\left(f_{\phi},f_{\tilde{\phi}}\right)\sim\mathcal{F}_{\mathcal{T}}}\left\Vert f_{\tilde{\phi}}-\mathcal{T}\left(f_{\phi}\right)\right\Vert ^{2}
}
\label{eq:mse}
\end{equation}

% \begin{align}
% l_{\tau} & =\mathbb{E}_{\left(f_{\phi},f_{\tilde{\phi}},f_{p}\right)\sim\mathcal{F}_{\mathcal{T}}}\left\Vert f_{\tilde{\phi}}-\mathcal{T}\left(f_{\phi},f_{p}\right)\right\Vert ^{2}\label{eq:mse}
% \end{align}
\noindent \textbf{Model implementation}. The functional $\mathcal{T}$
is implemented as a deep network, with the model architecture in Fig.~\ref{fig:The-pipeline-of}~(2).
It consists of a residual block, where the LeakyReLu activation function
is used to learn the nonlinear mapping from fully connection layers.
We employ BatchNorm and dropout to improve the generalisation of $\mathcal{T}$.
The skip connection is used to keep the scale of classifiers’ parameters and avoid the degradation
of learning. The pseudo-codes of sampling functional episodes and
MFL are shown in Alg.~\ref{alg:Meta-Functional-Learning}.

\begin{algorithm}[tb]
\caption{Meta Functional Learning (MFL). }
\label{alg:Meta-Functional-Learning}
\begin{algorithmic}[1] %[1] enables line numbers
\REQUIRE Embeddings $ \Psi_{src} = \{ \psi(\mathbf{I}_j), y_j \in \mathcal{C}_{base} \}$ of $D_{src}$; Classifier $f_c$; Sampling time $M_l$, $M_f$; Hyper-parameter set $H$; Shot number $s$, $s\times k$; Train epochs $T$;
\ENSURE Functional set $\mathcal{F}_{\mathcal{T}}$; Functional regularisation $\mathcal{T}$; 
\STATE // \textbf{Sampling Functional Episodes}
\STATE $\mathcal{F}_{\mathcal{T}}=\Phi$; $\mathcal{F}^{(b)}_{\mathcal{T}}=\Phi, b \in {\mathcal{C}_{base}}$;
\FOR{all $b\in \mathcal{C}_{base}$}
\STATE Sample episode $\mathcal{E}_{l}=\{\psi(\mathbf{I}_j^i),y_j=b\}_{i=1}^{N_b} \bigcup$ $\{ \psi(\mathbf{I}_j^i),y_j\neq b\}_{i=1}^{2 \times N_b}$ from $\Psi_{src}$ and train $f_{\tilde{\phi}}^{b}$ on $\mathcal{E}_{l}$;
\STATE Randomly sample sub-episode $\mathcal{E}_{f}$ including $s(s\times k$) $\psi(\mathcal{I}_j)$ with $y_j =(\neq) b$ from $\mathcal{E}_{l}$ and train $f_{\phi}^{(b)}$ on $\mathcal{E}_{f}$;
\STATE Compute $f_{p}^{(b)}$ including the prototypes of $\psi(\mathcal{I}_j)$ with $y_j=b$ and $y_j \neq b$ in $\mathcal{E}_{f}$;
\STATE $\mathcal{F}^{(b)}_{\mathcal{T}}=\mathcal{F}^{(b)}_{\mathcal{T}} \bigcup (f_{\phi}^{(b)},f_{\tilde{\phi}}^{(b)},f_{p}^{(b)})$;
\STATE Repeat line 6-7 using $f_c$ with $h$ in $H$;
\STATE Repeat line 5-8 for $M_f$ times;
\STATE Repeat line 4-9 for $M_l$ times;
\STATE $\mathcal{F}_{\mathcal{T}}=\mathcal{F}_{\mathcal{T}} \bigcup \mathcal{F}^{(b)}_{\mathcal{T}}$
\ENDFOR
\STATE // \textbf{Learning from the Scratch}
\WHILE {$t < T$}
\STATE Randomly split mini-batches with size $n$ from $\mathcal{F}_{\mathcal{T}}$;
\FOR {each mini-batch}
\STATE Predict functions $\mathcal{T}(f_{\phi}, f_{p})$ with $\mathcal{T}$;
\STATE Compute the loss in Eq.~(\ref{eq:mse});
\STATE Update the parameters of $\mathcal{T}$;
\ENDFOR
\ENDWHILE
\end{algorithmic}
\end{algorithm}

\subsection{ Generalised Forms of MFL }
We present a vanilla MFL method~\cite{Panijcai2021} by learning regularisation knowledge with the input of classifier's parameters for a binary classifier in Sec.~\ref{sec:vanilla-sampling} and Sec.~\ref{sec:vanilla-mfl}. 
To further exploit the potentialities of our MFL framework, we further propose  several generalised forms of MFL. Particularly,
the vanilla MFL is extended to learning from examples (Sec.~\ref{sub:examples}), multiple information source (Sec.~\ref{sub:multiple}) and with iterative updates (Sec.~\ref{sub:IU} ). And we further consider MFL in the wider applications: 1) learning functional for multi-class classifiers (Sec.~\ref{sub:multi-classifier}); 2) ensemble classifiers during inference phase (Sec.~\ref{sub:ensemble}).

% % \subsection{Generalizing  MFL}
% As the MFL and its variants learn regularisation knowledge for classifiers, they are not limited to the binary classifier. 
% %
% Here we extend them to wider applications: 1) learning functional for multi-classifier; 2) ensemble classifiers during inference phase.

% \subsubsection{Meta functional learning with prototypes (MFL-P)}

\subsubsection{Learning from the Examples \label{sub:examples}}
% Meta functional learning with prototypes (MFL-P)}

\begin{figure}
\centering{}
\includegraphics[scale=0.32]{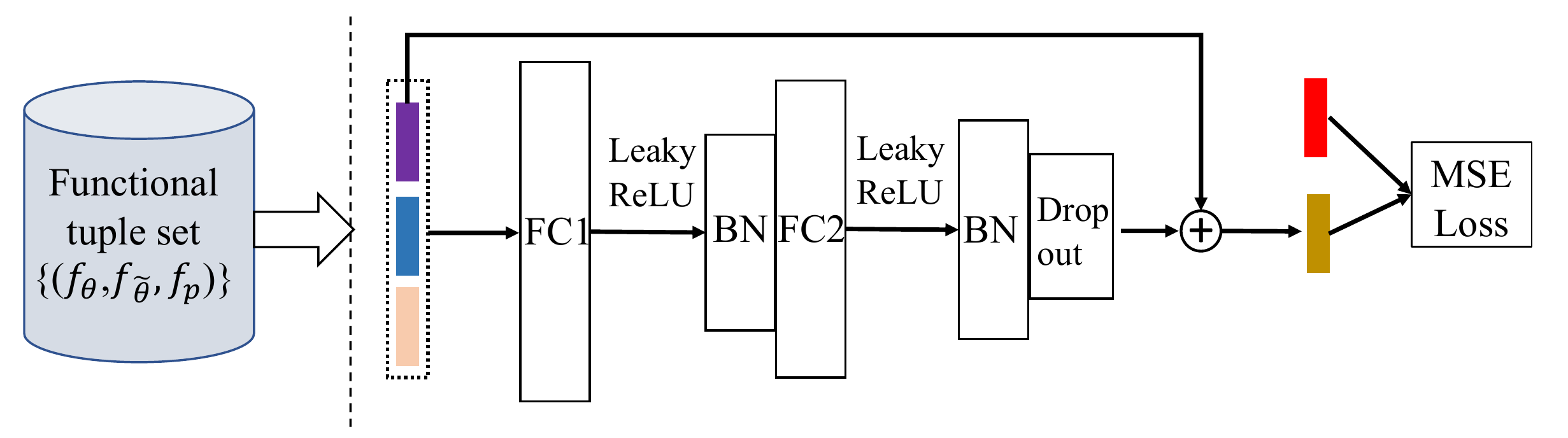}
\vspace{-0.15in}
\caption{The overall model design for Meta Functional Learning with Prototypes (MFL-P). The legend is same to Fig.~\ref{fig:The-pipeline-of}
\label{fig:The-pipeline-mfl-prototype}}
\end{figure}

In vanilla MFL, we learn functional $\mathcal{T}$ only using the classifier's parameter $f_{\phi}$.
The prototype $f_{p}$ from the representational space is ignored whilst it can provide important category-related information, to help $\mathcal{T}$ better learn the category agnostic knowledge in the meta training episodes.
Therefore, we improve the vanilla MFL by learning an extended form with prototypes (MFL-P), i.e. $\mathcal{T}:\left(f_{\phi},f_{p}\right)\rightarrow f_{\tilde{\phi}}$, where $f_{p}$ is a vector by concatenating the positive and negative prototypes, which are computed by averaging the embeddings of samples from corresponding classes.
% Given the functional sets $\mathcal{F}_{\mathcal{T}}$, we design
% a meta functional learning mechanism to learn the functional regularisation
% $\mathcal{T}$. For any given class $b$, 
In MFL-P, the objective is to approximate the ground-truth output $f_{\tilde{\phi}}^{\left(b\right)}=\mathcal{T}\left(f_{\phi}^{\left(b\right)},f_{p}^{\left(b\right)}\right)$.
We still use Mean Square Error (MSE) to measure the difference of
parameter vectors $\left(f_{\phi},f_{\tilde{\phi}},f_{p}\right)$
as, 
\begin{equation}
{
l_{\tau}=\mathbf{E}_{\left(f_{\phi},f_{\tilde{\phi}},f_{p}\right)\sim\mathcal{F}_{\mathcal{T}}}\left\Vert f_{\tilde{\phi}}-\mathcal{T}\left(f_{\phi},f_{p}\right)\right\Vert ^{2}
}
\label{eq:mse-mfl-p}
\end{equation}

% \subsubsection{Composite MFL (ComMFL)}
\subsubsection{Learning from Multiple Information Source\label{sub:multiple}}
\begin{figure}
\centering{}
\includegraphics[scale=0.32]{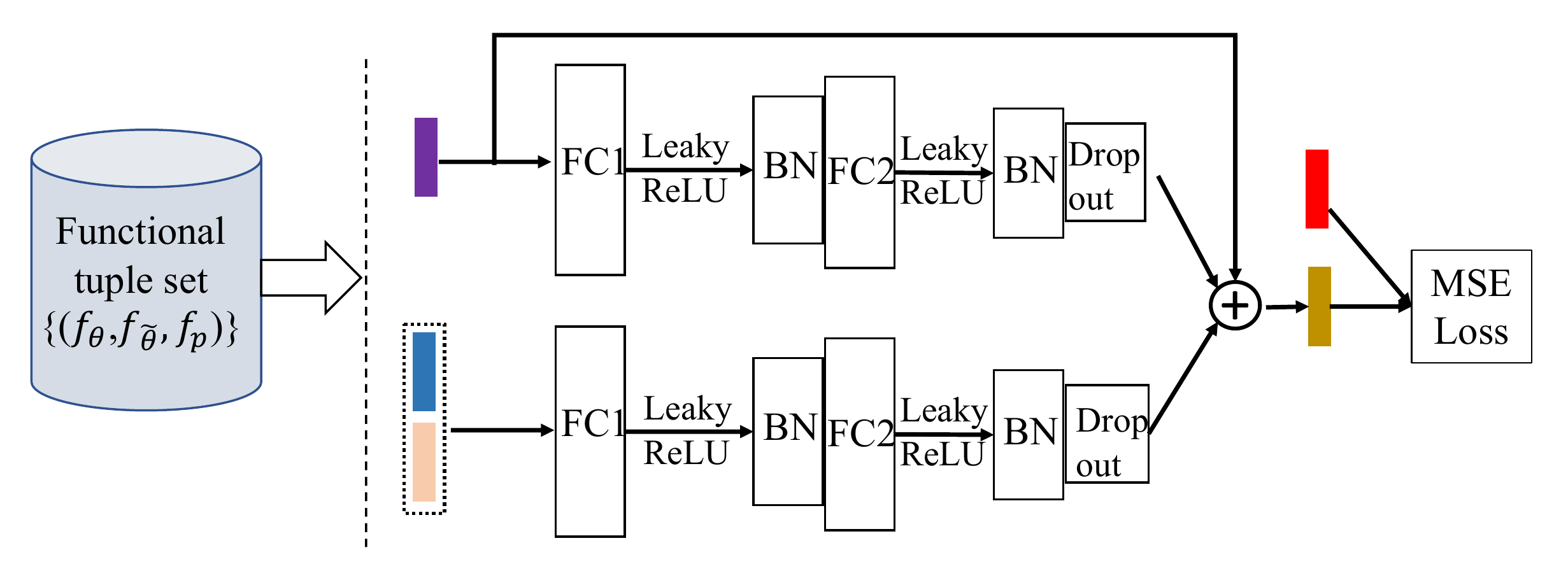}
\vspace{-0.1in}
\caption{The overall model design for Composite Meta Functional Learning (ComMFL). The legend is same to Fig.~\ref{fig:The-pipeline-of}
\label{fig:The-pipeline-ComMFL}}
\end{figure}
A vanilla MFL learns regularisation knowledge in the functional space and MFL-P further uses the prototypes as auxiliary knowledge for functional learning. However, these two types of MFL both focus on learning in the functional space whilst the classifier's function can also be learned from the representational space, that is, the prototypes. 
To learn better functions with comprehensive knowledge from both representational space and functional space, we propose a Composite MFL (ComMFL) by modifying the model of MFL-P. 
As in Fig.~\ref{fig:The-pipeline-ComMFL}, we use vanilla MFL to obtain a classifier's function by learning functional knowledge in functional space and additional a model to learn function in representational space. The objective function of ComMFL is formulated as:
\begin{equation}
{
l_{\tau}=\mathbf{E}_{\left(f_{\phi},f_{\tilde{\phi}},f_{\phi}\right)\sim\mathcal{F}_{\mathcal{T}}}\left\Vert f_{\tilde{\phi}}-( \mathcal{T}\left(f_{\phi}\right) +\mathcal{T}\left(f_{p}\right))\right\Vert ^{2}
}
\label{eq:mse-commfl}
\end{equation}

\subsubsection{MFL with Iterative Updates \label{sub:IU}} 
% \subsubsection{Incremental MFL (IMFL)} 
\begin{figure}
\centering{}
\includegraphics[scale=0.32]{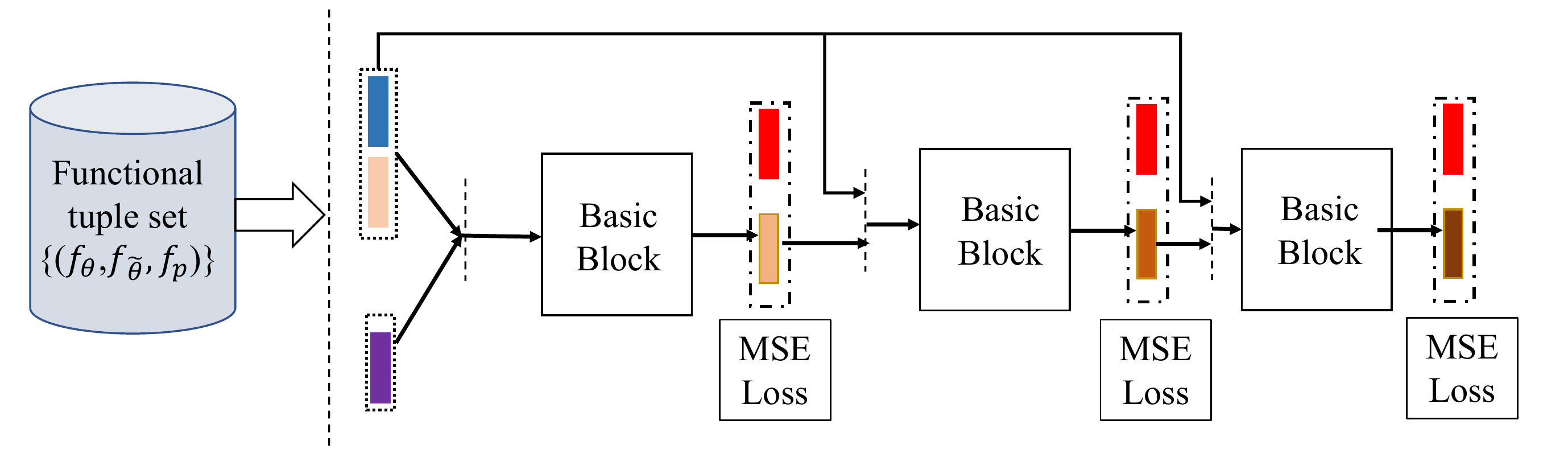}
\vspace{-0.25in}
\caption{The overall model design for Meta Functional Learning with Iterative Updates (MFL-IU). The legend is same to Fig.~\ref{fig:The-pipeline-of}
\label{fig:The-pipeline-mfl-iu}}
\end{figure}
The functionals in vanilla MFL, MFL-P and ComMFL are all optimised by a residual-based block with a MSE loss.
This one-step process with one module block may limit the capacity of functional to regularise FSL models, especially those trained with extremely-scarce labelled data.
Therefore, we employ an iterative update strategy on MFL (MFL-IU) to progressively learn the functional by a sequence of residual-based blocks.
Specifically, as illustrated in Fig.~\ref{fig:The-pipeline-mfl-iu},
MFL-IU$x$ has $x$ residual-based blocks and each block is optimised with a MSE loss. 
MFL-IU$x$ represents the output
of $x$th basic block, i.e. $\mathcal{T}_{x}\left(\mathcal{T}_{x-1}\cdots\left(\mathcal{T}_{1}\left(f_{\phi},f_{p}\right)\right)\right)$.
A simple version is MFL-IU$1$ by only using one block for vanilla MFL. The
training process of MFL-IU is illustrated in Alg.~\ref{alg:Meta-Functional-Learning-IE}.
Besides, we can employ this iterative update strategy on MFL-P and ComMFL, obtaining MFL-P-IU and ComMFL-IU.

\begin{algorithm}[t]
\begin{algorithmic}[1]
\REQUIRE Functional set $\mathcal{F}_{\mathcal{T}}$; Iterations $X$; Train epochs $T$;
\ENSURE Functional regularisation $\mathcal{T}=\{\mathcal{T}_1,...,\mathcal{T}_X\}$; 
\WHILE {$t < T$}
\STATE Randomly split mini-batches with size $n$ from $\mathcal{F}_{\mathcal{T}}$;
\FOR {each mini-batch}
\FOR {$x < X$}
\STATE Predict functions $\mathcal{T}_x(f_{\phi}, f_{p})$ with $\mathcal{T}_x$;
\STATE Compute the loss in Eq.~(\ref{eq:mse});
\STATE Update the parameters of $\mathcal{T}_x$;
\ENDFOR
\ENDFOR
\ENDWHILE
\end{algorithmic}
\caption{MFL with Iterative Updates. \label{alg:Meta-Functional-Learning-IE}}
\end{algorithm}

\pan{
\subsubsection{MFL on Multi-class Classifiers\label{sub:multi-classifier}}
For a binary classifier, the task-agnostic knowledge learned by functional is simplified as a learning problem at a category-level. That is, assuming that the learned regularisation knowledge is category-agnostic and can be transferred across classes.
In this way, the functional set are sampled according to different positive classes, and its scale linearly increases with the number of classes in source dataset.
However, for multi-class classifiers, the functional should be capable of capturing the regularisation knowledge of $N$ different classes, as well as their relationships.
Obviously, learning MFL for a multi-class classifier is a more challenging task than that for a binary classifier.}

\pan{
To solve the increased complexity in the functional learning for a multi-class classifier, we extend our MFLs to this scenario by sampling more tuples in a functional set. 
Specifically, we adopt an outer loop strategy on the MFLs for a binary classifier, and the inner loop is a complete training of MFLs.
% Specifically, for sampling functional set for $N$-way multi-classifier from source dataset with $M$ categories, the task's complexity  $A_M^N$ combinations and compute a set of tuples for each combination.
%
We train inner loop $I_{out}$ times and the algorithm is detailed in Alg.~\ref{alg:MFL-multi-classifier}.
By this way, we learn the functional capturing more tasks while avoiding the excessive increase of the storage cost for a functional set. 
}
\begin{algorithm}[t]
\begin{algorithmic}[1]
\REQUIRE Train epochs $T$; Outer loop time: $I_{out}$.
\ENSURE Functional regularisation $\mathcal{T}=\{\mathcal{T}_1,...,\mathcal{T}_X\}$;
\FOR {$i<I_{out}$}
\STATE \textbf{Sampling functional episodes $\mathcal{F}_{\mathcal{T}}$};
\WHILE {$t < T$}
\STATE Train MFLs on $\mathcal{F}_{\mathcal{T}}$;
\ENDWHILE
\ENDFOR
\end{algorithmic}
\caption{MFL on Multi-class Classifiers. \label{alg:MFL-multi-classifier}}
\end{algorithm}

\pan{
\subsubsection{MFL as an Ensemble of Classifiers\label{sub:ensemble}}
Ensemble method is a machine learning technique that combines several base models in order to produce one optimal predictive model.
Generally, the simple ensemble methods, e.g. average the weights or prediction scores of every base-classifier, prefer to yield a moderate prediction results compared to the base-classifiers.
However, inferior base-classifier might introduce noisy predictions, resulting negative affect on the ensemble model.
We introduce a MFL method that can improve both the base-classifiers and further benefit the ensemble results.
Specifically, during the training phase, the MFL gradually captures the converges behaviour of classifiers trained with different hyper-parameters since we sample them into functional set. 
Thus, this MFL can be used as an ensemble of classifiers with different hyper-parameters; especially for the hyper-parameter-sensitive classifier trained with limited data.
So the integrated classifier by MFL can be formulated as $$f_{\mathcal{T}}^{ens}=1/C\sum_{c=1}^{C}\lambda_{c} \mathcal{T}(f_c),$$
where $\mathcal{T}$ is the MFL module, $f_c$ is the classifier trained on limited data with hyper-parameters $c$ and $\lambda_c$ is the corresponding weight for $\mathcal{T}(f_c)$.
In this work, we simply use $\lambda_k$ as 1 for every classifier.
}

\section{Experiments}
To evaluate the effectiveness of MFL, we tested MFL on two data-scarce
learning problems: $N$-way $K$-shot classification, i.e. a task
aiming to discriminate between $N$ classes with $K$ labelled samples
of each class, by (1) standard FSL and (2) Cross-Domain FSL (CD-FSL).
In particular, we adopted a binary classifier as a vanilla classifier and generalised
it to multi-way classification scenario with one\emph{ vs.}all manner. We first
evaluated MFL on basic 2-way FSL tasks and then investigated whether
the learning pattern of MFL can be generalised to multi-way FSL
tasks.
\pan{
We also evaluate our MFL on multi-class classifiers for corresponding multi-way FSL tasks.
}
Furthermore, the experiments on CD-FSL were carried out for
learning tasks with different shot numbers to investigate the model generalisation
capacity to multi-shot FSL tasks.

\noindent \textbf{Datasets.} We employed three FSL datasets: 1){\emph{mini}ImageNet} is a subset of the ILSVRC-12~\cite{russakovsky2015imagenet} dataset and contains
100 classes with 600 images per class. We followed the split in~\cite{ravi2016optimization}
and used 64, 16 and 20 classes as base, validation and novel sets, respectively.
2) {{\emph{CIFAR-FS}}} is a dataset with lower-resolution images, and it contains
100 classes with 600 instances in each class. Following the split
in ~\cite{bertinetto2019meta}, we used 64 classes to construct the
base set, 16 and 20 for validation and novel set.
3) \emph{{CUB}} is a fine-grained dataset which consists of 200 bird categories with 11788
images in total. We used 100, 50 and 50 classes for base, validation
and novel sets with the previous setting in~\cite{hilliard2018few},
and we conducted all experiments with the cropped images provided
in~\cite{triantafillou2017few}.
\pan{ 
4) \emph{{Cars}}~\cite{KrauseStarkDengFei-Fei_cars} contains 16,185 images of 196 classes of cars. We follow the split in~\cite{Tseng2020Cross-Domain} and used 98, 49 and 49 classes as base, validation and novel sets.
5) \emph{{Places}}~\cite{zhou2017places} is a dataset for scene recognition with 365 categories and 8 millions of images. We used 183, 91 and 91 classes as base, validation and novel sets following the split in~\cite{Tseng2020Cross-Domain}.
}

\begin{table}[t]
\caption{Hyper-parameters for sampling functional episodes. \#way represents the number of classes in multi-class classifier.} \label{tab:paras-functional_episodes}
\centering 
% \resizebox{0.9\columnwidth}{!}
% {
\begin{tabular}{lcc}
\toprule
& Binary classifier & multi-class classifier  \\
\midrule
\#Outer loop       & 1            & 5  \\
\#Many-shot model $M_l$ & 5*64         & 500  \\
\#Few-shot model  $M_s$ & 100          & 200  \\
Negative samples $k$    & $\{1,2,3,4\}$  &\#way-1  \\
Hyper-parameter set $H$ & $1e\{-2,-1,0,1,2\}$  & $1e\{-2,-1,0,1,2\}$  \\
\#Functional episodes   & 5*64*100*5*5 & 5*500*200*1*5 \\
\bottomrule
\end{tabular}
% }
\end{table}

\noindent \textbf{Implementation. }
We used Conv4 as the backbone for learning a feature representation.
The architecture of this Conv4 network is provided by~\cite{snell2017prototypical} and it contains four convolutional blocks.
Each block comprises a 64-filter 3 \texttimes{} 3 convolution, batch normalization layer, a ReLU nonlinearity and a 2 \texttimes{} 2 max-pooling layer. 
%'
For training the representator, we randomly split the images from base classes into (90\%, 10\%) partition as (train, validation) sets. We trained the backbone over 120 epochs.
We use SGD optimizer with a momentum of 0.9 and a weight decay of 1$e-$4. 
We set batch size as 64 and the learning rate is initialized as
0.01 and decayed with a factor of 0.1 by three times.
For training MFL and its variants, we employed BatchNorm (0.1), dropout (0.9) and LeakyReLU (0.01), and the parameters for the first and second fully connected layers are 600 and 1601 respectively.
Moreover, we trained MFL and its variants over 30 epochs with batch size (256), and the learning rate is initialised as 0.01 and decay to 1$e$-3 after 20 epochs.
We adopted the Logistic Regression (LR) function as a base binary classifier or multi-class classifier.
The hyper-parameters for sampling functional episodes for binary classifiers and multi-class classifiers are shown in Tab.~\ref{tab:paras-functional_episodes}.
Specifically, we set $s=\{1,2,3,4,5\}$ to construct functional tuple sets for $s$-shot learning scenarios in FSL.
% and the parameters for computing functional set are $M_{l}=5,M_{s}=100,$
% $k=\{1,2,3,4\}$ and $H=1e\{-2,-1,0,1,2\}$. 
% %
% Specifically, we set $s=\{1,2,3,4,5\}$ to construct functional tuple sets for $s$-shot learning scenarios in FSL.
% %
% For multi-classifier, we set $M_{l}=500,M_{s}=100$.
%
In all experiments, we selected the best
model by evaluating them on a validation set and evaluated all methods
with 600 episodes randomly selected from the novel classes in the
corresponding dataset.

\subsection{Meta Functional Learning}
\subsubsection{MFL for Binary Classifier}
\label{sub:var-MFL}

\begin{table*}[t]
\caption{Few-Shot Learning Evaluation: Comparison to Vanilla LR and prior work on \emph{mini}ImageNet and \emph{CIFAR-FS} with Conv4 backbone. Mean accuracies (\%) with 95\% confidence intervals
results are reported on $N$-way 1-shot FSL. $(\cdot)^\dagger$ represent the experimental results with the released
codes and $(\cdot)^\ddagger$ are our re-implemented results with the corresponding paper. 
\textbf{Bold:} the best scores.
}
\label{tab:Mean-Accuracies-of-maintable}
\centering
% \resizebox{1.8\columnwidth}{!}
% {
\begin{tabular}{llcccccc}
\toprule 
\multirow{1}{*}{Dataset} & \multirow{1}{*}{Methods} & 2-way & 3-way & 4-way & 5-way & 10-way & 20-way \tabularnewline
\midrule
\multirow{9}{*}{\textbf{\emph{mini}ImageNet}} 
& Baseline$^{\dagger}$~\cite{chen2019closer} & {\small{}70.09$\pm$1.13} & {\small{}55.74$\pm$0.99} & {\small{}46.33$\pm$0.79} & {\small{}40.41$\pm$0.68} & {\small{}26.50$\pm$0.38} & {\small{}16.09$\pm$0.21}\tabularnewline
 & ProtoNet$^{\dagger}$~\cite{snell2017prototypical} & {\small{}73.76$\pm$1.34} & {\small{}59.34$\pm$1.14} & {\small{}51.24$\pm$0.95} & {\small{}45.22$\pm$0.81} & {\small{}29.04$\pm$0.44} & {\small{}18.09$\pm$0.23}\tabularnewline
 & MAML$^{\dagger}$~\cite{finn2017model} & {\small{}73.56$\pm$1.38} & {\small{}62.21$\pm$1.16} & {\small{}52.44$\pm$0.94} & {\small{}48.29$\pm$0.83} & {\small{}31.41$\pm$0.47} & {\small{}-}\tabularnewline
\cmidrule{2-8} 
 & Vanilla LR & {\small{}72.86$\pm$1.13} & {\small{}59.51$\pm$0.93} & {\small{}51.05$\pm$0.83} & {\small{}46.18$\pm$0.77} & {\small{}31.04$\pm$0.44} & {\small{}21.09$\pm$0.24}\tabularnewline
 & MetaModelNet$^{\ddagger}$~\cite{wang2017learning} & {\small{}76.34$\pm$1.36} & {\small{}62.54$\pm$1.14} & {\small{}53.51$\pm$0.97} & {\small{}47.99$\pm$0.85} & {\small{}31.02$\pm$0.46} & {\small{}19.23$\pm$0.24}\tabularnewline
 \cmidrule{2-8} 
 & vanilla MFL (Ours) & {\small{}76.09$\pm$1.15} & {\small{}62.70$\pm$1.00} & {\small{}54.37$\pm$0.86} & {\small{}48.88$\pm$0.80} & {\small{}33.15$\pm$0.46} & {\small{}22.42$\pm$0.25} \\
%  & MFL-P (Ours) & {\small{}76.63$\pm$1.16} & {\small{}63.19$\pm$0.98} & {\small{}54.90$\pm$0.86} & {\small{}49.10$\pm$0.80} & {\small{}33.36$\pm$0.45} & {\small{}22.54$\pm$0.25}\tabularnewline
%  & ComMFL-P (Ours) & {\small{}76.96$\pm$1.19} & {\small{}63.87$\pm$1.01} & {\small{}55.14$\pm$0.87} & {\small{}49.65$\pm$0.80} & {\small{}33.71$\pm$0.45} & {\small{}22.75$\pm$0.26}\tabularnewline
% \cmidrule{2-8} 
 & vanilla MFL-IU3 (Ours) & {\small{}77.60$\pm$1.23} & {\small{}64.62$\pm$1.03} & {\small{}56.40$\pm$0.88} & {\small{}50.87$\pm$0.82} & {\small{}34.43$\pm$0.45} & {\small{}23.22$\pm$0.27}\tabularnewline
 & MFL-P-IU3 (Ours) & {\small{}78.41$\pm$1.21} & {\small{}65.47$\pm$1.02} & {\small{}56.77$\pm$0.90} & {\small{}51.46$\pm$0.83} & {\small{}34.88$\pm$0.46} & {\small{}23.64$\pm$0.26}\tabularnewline
 & ComMFL-IU3 (Ours) & \textbf{\small{}78.83$\pm$1.23} & \textbf{\small{}65.90$\pm$1.03} & \textbf{\small{}57.56$\pm$0.92} & \textbf{\small{}52.03$\pm$0.83} & \textbf{\small{}35.27$\pm$0.46} & \textbf{\small{}23.72$\pm$0.23}\tabularnewline
\midrule
\multirow{9}{*}{\textbf{\emph{CIFAR-FS}}} 
& Baseline$^{\dagger}$~\cite{chen2019closer} & {\small{}72.66$\pm$1.14} & {\small{}59.44$\pm$1.06} & {\small{}50.77$\pm$0.85} & {\small{}46.16$\pm$0.77} & {\small{}32.46$\pm$0.46} & {\small{}22.04$\pm$0.26}\tabularnewline
 & ProtoNet$^{\dagger}$~\cite{snell2017prototypical} & {\small{}73.36$\pm$1.13} & {\small{}60.45$\pm$1.20} & {\small{}51.87$\pm$1.01} & {\small{}47.04$\pm$0.91} & {\small{}31.41$\pm$0.51} & {\small{}20.48$\pm$0.25}\tabularnewline
 & MAML$^{\dagger}$~\cite{finn2017model} & {\small{}75.82$\pm$1.35} & {\small{}63.06$\pm$1.23} & {\small{}56.82$\pm$1.03} & {\small{}50.15$\pm$0.94} & {\small{}39.52$\pm$0.60} & {\small{}-}\tabularnewline
\cmidrule{2-8} 
 & Vanilla LR & {\small{}76.53$\pm$1.16} & {\small{}64.12$\pm$1.02} & {\small{}56.62$\pm$0.92} & {\small{}51.48$\pm$0.82} & {\small{}38.67$\pm$0.49} & {\small{}28.27$\pm$0.28}\tabularnewline
 & MetaModelNet$^{\ddagger}$~\cite{wang2017learning} & {\small{}79.37$\pm$1.25} & {\small{}67.96$\pm$1.23} & {\small{}60.11$\pm$1.11} & {\small{}55.26$\pm$1.02} & {\small{}39.48$\pm$0.61} & {\small{}27.09$\pm$0.31}\tabularnewline
\cmidrule{2-8} 
 & vanilla MFL (Ours) & {\small{}80.11$\pm$1.14} & {\small{}68.99$\pm$1.03} & {\small{}61.10$\pm$0.96} & {\small{}55.90$\pm$0.88} & {\small{}42.35$\pm$0.53} & {\small{}30.62$\pm$0.28}\tabularnewline
%  & MFL-P (Ours) & {\small{}81.09$\pm$1.13} & {\small{}69.75$\pm$1.06} & {\small{}62.06$\pm$0.96} & {\small{}57.18$\pm$0.88} & {\small{}42.73$\pm$0.53} & {\small{}31.38$\pm$0.29}\tabularnewline
%  & ComMFL-P (Ours) & {\small{}80.86$\pm$1.17} & {\small{}70.48$\pm$1.06} & {\small{}62.70$\pm$0.97} & {\small{}57.72$\pm$0.90} & {\small{}43.63$\pm$0.55} & {\small{}31.56$\pm$0.29}\tabularnewline
%  \cmidrule{2-8}
 & vanilla MFL-IU3 (Ours) & {\small{}81.39$\pm$1.17} & {\small{}71.60$\pm$1.09} & {\small{}63.88$\pm$0.99} & {\small{}59.38$\pm$0.93} & {\small{}45.25$\pm$0.58} & {\small{}32.78$\pm$0.29}\tabularnewline
 & MFL-P-IU3 (Ours) & \textbf{\small{}82.68$\pm$1.12} & {\small{}72.37$\pm$1.08} & {\small{}64.71$\pm$0.99} & {\small{}59.88$\pm$0.93} & {\small{}45.25$\pm$0.58} & {\small{}32.78$\pm$0.29}\tabularnewline
 & ComMFL-IU3 (Ours) & {\small{}82.64$\pm$1.18} & \textbf{\small{}72.83$\pm$1.10} & \textbf{\small{}65.15$\pm$1.02} & \textbf{\small{}60.33$\pm$0.94} & \textbf{\small{}45.67$\pm$0.59} & \textbf{\small{}33.31$\pm$0.29}\tabularnewline
\bottomrule
\end{tabular}
% }
\end{table*}

\noindent \textbf{Competitors.}
We compared our methods against existing models for $N$-way 1-shot FSL tasks from three perspectives: 1) Comparison with the base classifier: We used Logistic Regression (LR) as a typical classifier.
As in Tab.~\ref{tab:Mean-Accuracies-of-maintable}, the Vanilla LR represents a naive LR classifier trained on labelled data, while Vinilla MFL, Vinilla MFL-IU$3$, MFL-P-IU3 and ComMFL-IU$3$ are the predicted functions with corresponding models.
2) Comparison with typical FSL methods: Baseline~\cite{chen2019closer}
ProtoNet~\cite{snell2017prototypical}, and MAML~\cite{finn2017model};
3) Comparison with a model transformation method: MetaModelNet~\cite{wang2017learning}.
Since no official results are provided on these comparison methods in $N$-way classification FSL, we re-ran the released code in~\cite{chen2019closer} for evaluating existing FSL methods and evaluated MetaModelNet with our re-implemented model following~\cite{wang2017learning}.

% \noindent \textbf{Complexities Analysis.} 
% The complexities of the proposed MFL method and the competitors are: 1)ProtoNet is the simplest one which uses some episodes to train backbone; 2)Baseline is simple but needs training backbone with 400 epochs; 3) MAML has higher complexity since it needs to compute high-order gradient; 4) Our MFL methods only need to compute one-order gradient.

\noindent \textbf{Results and analysis.} Table~\ref{tab:Mean-Accuracies-of-maintable} shows the comparative results on \emph{mini}ImageNet and \emph{CIFAR-FS}.
We can see that: (1) Our methods can effectively transfer the regularisation knowledge to benefit the naive functions, i.e. Vanilla LR, yielding more robust and accurate functions with significant performance improvement
on 2/3/4/5/10/20-way 1-shot FSL;
(2) Our methods significantly outperform
three typical FSL methods, achieving the potentially smooth and discriminative hypotheses on a fixed embedding space; 
(3) MetaModelNet can improve the performance of the Vanilla LR in low-way (1-5 way) FSL tasks, while the improvement in higher way (10/20 way) FSL tasks
is limited. In contrast, our methods performed well in all $N$-way
1-shot FSL tasks. This verifies that our methods on binary classifiers are more robust and generalisable to multi-way FSL tasks. 

\noindent \pan{\textbf{Effects of generalised forms of MFL. }
In Tab.~\ref{tab:Mean-Accuracies-of-maintable}, we observe that all forms of MFL, i.e vanilla MFL, vanilla MFL-IU3, MFL-P-IU3 and ComMFL-IU3, are effective in improving the performance of the Vanilla LR. 
In particular, vanilla MFL-IU3 performed better than vanilla MFL
due to the benefit from the progressively increasing functional regularisation
knowledge provided by the iterative update strategy.
Essentially, involving the information of examples can benefit the functional learning of regularisation knowledge.
As expected, the results of MFL-P-IU3 and ComMFL-IU3 show better performance on $N$-way 1-shot FSL tasks compared with vanilla MFL-IU3.
We note that the two ways to explore the information from examples perform slightly differently. That is, ComMFL-IU3 obtains slightly better performance than MFL-P-IU3, suggesting that the composition of different functionals, i.e. a functional from examples in the representational space and a functional from functions in the functional space, is a better choice to improve the learning of generalisable regularisation knowledge.
}

\subsubsection{MFL for Multi-class Classifiers}
\pan{
% \noindent \textbf{MFL for multi-classifiers. } 
In~\ref{sub:var-MFL}, we present extensive experimental results to verify the effectiveness of the various forms of MFL for improving a binary classifier with few labelled data.
To further evaluate the generalisation ability of our methods on a multi-class classifier, we conducted experiments on 3/4/5-way 1-shot FSL tasks by learning functional regularisation on corresponding 3/4/5-class classifiers. 
In particular, the hyper-parameters for sampling functional episodes are in Tab.~\ref{tab:paras-functional_episodes}. 
Note that the number of functional episodes for a multi-class classifier is larger than that for a binary classifier to satisfies the requirements of larger functional space.
%
% For the episodes of 3/4/5-way 1-shot task, we compute functions with corresponding multi-class LR.
\begin{table}[!ht]
\caption{Multi-class classifier evaluation: Mean accuracies (\%) of Vanilla LR and LR with MFL and MFL-IU3 on 3/4/5-way 1-shot tasks from \emph{mini}ImageNet.
\textbf{Bold:} the best scores.}
\label{tab:multi-classifier}
\centering
% {}\centering {\small{}{}}%
\begin{tabular}{lccc}
\toprule 
{\small{}{}\#shot} & {\small{}{}3-way}& {\small{}{}4-way} & {\small{}{}5-way} \tabularnewline
\midrule 
{\small{}{}Vanilla LR} & {\small{}59.69$\pm$0.93} & {\small{}51.16$\pm$0.83} & {\small{}46.22$\pm$0.77} \tabularnewline
{\small{}{}vanilla MFL} & {\small{}61.72$\pm$0.99} & {\small{}53.01$\pm$0.85} & {\small{}47.82$\pm$0.77} \tabularnewline
{\small{}vanilla MFL-IU3} & \textbf{\small{}62.97$\pm$1.00} & {\small{}54.34$\pm$0.85} & {\small{}48.72$\pm$0.75} \tabularnewline
{\small{}{}MFL-P-IU3} & {\small{}61.85$\pm$0.97} & {\small{}53.48$\pm$0.82} & {\small{}47.82$\pm$0.78}\tabularnewline
{\small{}{}ComMFL-IU3}& \textbf{\small{}63.00$\pm$0.98} & \textbf{\small{}54.63$\pm$0.85} & \textbf{\small{}48.88$\pm$0.77} \tabularnewline
\bottomrule
\end{tabular}
\end{table}
}

\noindent \pan {\textbf{Binary classifier $vs.$ Multi-class classifier. }
As we illustrated in Sec.~\ref{sub:multi-classifier}, MFL for multi-class classifiers is more challenging due to the functional space for multi-class classifiers is larger and hard to capture.
The results in Tab.~\ref{tab:Mean-Accuracies-of-maintable} and Tab.~\ref{tab:multi-classifier} valid this assumption and we observe that the functional learning on a binary classifier is more effective than that on a multi-class classifier.
In particular, for the 1-shot 5-way FSL tasks, the ComMFL-IU3 on a binary classifier obtains 52.03\% whilst that on a multi-class classifier get an inferior result 48.88\%,
and this observation is similar in the 3/4-way 1-shot FSL tasks.
Interestingly, with an auxiliary information from examples, MFL-P-IU3 performs inferior to vanilla MFL-IU3.
This observation is reverse to the results on binary classifiers,
which is counterintuitive and indicates that the samples might guide a biased learning for the regularisation functional on the functional space for a multi-class classifier.
Additionally, ComMFL-IU3 and vanilla MFL-IU3 obtain competitive results on a multi-class classifier.
This benefits from the individually networks to learn the functional on the functional space and the representational space, such the examples would not directly affect the functional learning on the functional space.   
}

\subsubsection{MFL as an Ensemble of Classifiers}
% \noindent \textbf{MFL as an ensemble of classifiers. }
\pan{
We conducted a simply average strategy on the predicted functions by the MFL-regularised classifiers using different hyper-parameters, i.e. $C=0.1, 1, 10$, and we compute a more accurate functions compared with each classifier with MFL.
The favour of ensemble method is preferring to yields a moderate results compared to best base-classifier, and this also occurs in the few-shot learning tasks shown in the Tab.~\ref{tab:Mean-Accuracies-ensemble} for the ensemble of Vanilla LR. 
With a weight averaging strategy, the ensemble method performs competitively well compared with the best base-classifier, achieving the same recognition result (46.18\%).
We also use this weight averaging strategy to integrate the weights of functions predicted by our MFLs.
Table~\ref{tab:Mean-Accuracies-ensemble} shows that the averaged results on the functions predicted by MFL perform better than those of each base-classifiers. This suggests that  it is a good choice of using ensemble methods after our MFLs.
We conjecture that our MFLs can transform the inferior classifiers trained with limited labels to more accurate ones, so that the ensemble method on the transformed classifiers can compute a more robust classifier and remitting the negative effects from the inferior classifiers without MFLs.
}

\begin{table}[!t]
\caption{MFL as an ensemble: Mean accuracies (\%) of our methods with Conv4 backbone on 5-way 1-shot tasks from \emph{mini}ImageNet.
\textbf{Bold:} the best scores. \underline{Underline:} the secondary best scores.}
\label{tab:Mean-Accuracies-ensemble}
\centering 
\resizebox{0.98\columnwidth}{!}
{
\begin{tabular}{lcccc}
\toprule
$\#C$      & 0.1 & 1.0 & 10  & Weight Ave. \\
\midrule
Vanilla LR   &  45.87$\pm$0.77 & 46.13$\pm$0.77 & \textbf{46.18$\pm$0.77}  &\textbf{46.18$\pm$0.77}  \\
vanilla MFL  &  \underline{49.60$\pm$0.76} & \textbf{49.94$\pm$0.78} & 48.88$\pm$0.80  &49.33$\pm$0.79  \\
vanilla MFL-IU3 & 49.35$\pm$0.77 & 50.50$\pm$0.78 & \underline{50.98$\pm$0.81}  & \textbf{51.11$\pm$0.81}     \\
MFL-P-IU3    & 50.75$\pm$0.84 & \underline{
51.94$\pm$0.82} & 51.48$\pm$0.81 & \textbf{52.33$\pm$0.77} \\
ComMFL-IU3   &  51.13$\pm$0.85 & \underline{52.06$\pm$0.85} & 52.03$\pm$0.83  & \textbf{52.37$\pm$0.84} \\
\bottomrule
\end{tabular}
}
\end{table}

\subsection{Learning to Cross Domain}
We employed our MFL methods on a more challenging task, CD-FSL. 
We followed the \emph{mini}ImageNet $\rightarrow$ CUB setting in~\cite{chen2019closer},
where $D_{src}$ and $D_{nov}$ are the images from the base classes of \emph{mini}ImageNet
and the novel classes of CUB, respectively.
Moreover, we generalise this setting to another two datasets, i.e. Cars and Places.
For comparison, we adopted the same competitors 
in Sec.~\ref{sub:var-MFL} and carried
out experiments on CD-FSL by using 5-way 1/5-shot settings referring to~\cite{chen2019closer}.
% to investigate model effectiveness on different shot learning scenarios.

\noindent \pan{\textbf{Analysis. }
Table~\ref{tab:Mean-Accuracies-of-CD} shows the results with the
following observations: (1) By directly using the learned representation
trained on \emph{mini}ImageNet, the three existing FSL methods give inferior
performance on CD-FSL. (2) MetaModelNet, the model transformation
method, improved the Vanilla LR on FSL but failed on CD-FSL, resulting
in a poorer transformed classifier than Vanilla LR. (3) Our methods
are able to improve the Vanilla LR by transferring the regularisation
knowledge in model learning across domains, yielding a more accurate classifier with 1\%-3\% increase of classification accuracy on 5-way $K$-shot CD-FSL under scenarios of mini$\rightarrow$CUB and mini$\rightarrow$Places.
Additionally, we note the improvement on mini$\rightarrow$Cars is limited, which might be due to that the embedding space pre-trained on \emph{mini}ImageNet is less-discriminative for Cars, such the assumption of continuity, cluster and manifold distributions for regularisation knowledge transfer is less effective.  
}
\begin{table*}[!ht]
\caption{Cross-Domain Few-Shot Learning Evaluation: Mean accuracies (\%) of our methods and the competitors with Conv4 backbone on 5-way $K$-shot tasks under the cross-domain scenarios.
\textbf{Bold:} the best scores.
}
\label{tab:Mean-Accuracies-of-CD}
\centering
% {}\centering {\small{}{}}%
% \resizebox{1.6\columnwidth}{!}
% {
\begin{tabular}{lcccccc}
\toprule 
{\small{}{}Dataset} & \multicolumn{2}{c}{\textbf{\small{}{}mini $\rightarrow$ CUB}} & \multicolumn{2}{c}{\textbf{\small{}{}mini $\rightarrow$ Cars}} & \multicolumn{2}{c}{\textbf{\small{}{}mini $\rightarrow$ Places}}\tabularnewline
% \midrule 
{\small{}{}\#shot} & 1 & 5 & 1 & 5 & 1 & 5\tabularnewline
\midrule 
Baseline~\cite{chen2019closer} & {\small{}36.57$\pm$0.57} &{\small{}58.74$\pm$0.69} & {\small{}26.41$\pm$0.57} &{\small{}36.24$\pm$0.59} & {\small{}39.64$\pm$0.69} &{\small{}60.80$\pm$0.74} \tabularnewline
ProtoNet~\cite{snell2017prototypical} & {\small{}42.00$\pm$0.74} & {\small{}64.24$\pm$0.70} & {\small{}28.53$\pm$0.57} & {\small{}41.78$\pm$0.69} & {\small{}41.13$\pm$0.75}&{\small{}63.07$\pm$0.72} \tabularnewline
MAML~\cite{finn2017model} & {\small{}39.87$\pm$0.69} & {\small{}58.26$\pm$0.76} & {\small{}29.36$\pm$0.61} & {\small{}37.12$\pm$0.63} & {\small{}44.46$\pm$0.80}&{\small{}52.87$\pm$0.75} \tabularnewline
\midrule 
{Vanilla LR} & {\small{}42.67$\pm$0.69} & {\small{}66.07$\pm$0.70} & {\small{}29.70$\pm$0.53} & {\small{}42.01$\pm$0.66} & {\small{}46.69$\pm$0.77}&{\small{}64.56$\pm$0.71} \tabularnewline
MetaModelNet~\cite{wang2017learning} & {\small{}36.57$\pm$0.76} & {\small{}52.73$\pm$0.78} & {\small{}25.38$\pm$0.49} & {\small{}31.22$\pm$0.52} & {\small{}43.61$\pm$0.86}& {\small{} 59.14$\pm$0.79 } \tabularnewline
% \midrule
{vanilla MFL (Ours)} & {\small{}44.48$\pm$0.71} & {\small{}67.20$\pm$0.70} & {\small{}29.99$\pm$0.53} & {\small{}42.19$\pm$0.68} & {\small{}47.95$\pm$0.83}&{\small{}65.33$\pm$0.72} \tabularnewline
{vanilla MFL-IU3 (Ours)} & {\small{}45.19$\pm$0.73} & {\small{}67.80$\pm$0.70} & {\small{}29.93$\pm$0.52} & {\small{}41.65$\pm$0.66} & {\small{}48.82$\pm$0.85}&{\small{}65.92$\pm$0.71} \tabularnewline
{MFL-P-IU3 (Ours)} & {\small{}45.37$\pm$0.77} & {\small{}67.85$\pm$0.71} & \textbf{\small{}30.26$\pm$0.54} & \textbf{\small{}42.31$\pm$0.68} & \textbf{\small{}49.71$\pm$0.85}&{\small{}66.61$\pm$0.69} \tabularnewline
{ComMFL-IU3 (Ours)} & \textbf{\small{}46.56$\pm$0.78} & \textbf{\small{}68.31$\pm$0.70} & {\small{}30.05$\pm$0.56} & {\small{}42.22$\pm$0.68} & {\small{}49.68$\pm$0.85} &\textbf{\small{}66.91$\pm$0.70} \tabularnewline
\bottomrule
\end{tabular}
% }
\end{table*}

\subsection{Ablation Study}
\begin{figure}[!ht]
\centering
\includegraphics[scale=0.9]{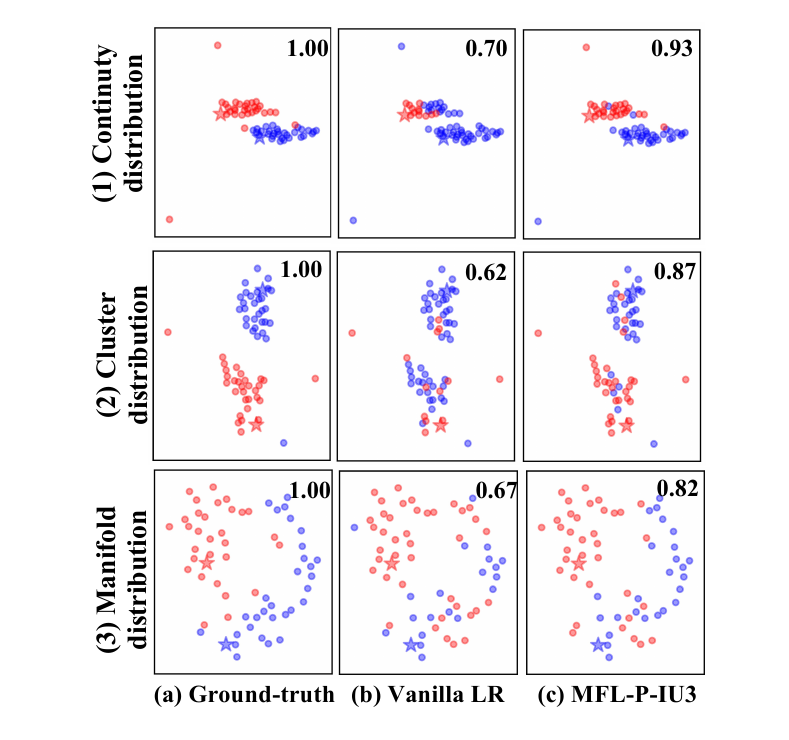}
\vspace{-0.2in}
\caption{The T-SNE visualisation of 2-way 1-shot FSL tasks from \emph{mini}ImageNet.
Plots~(a) depict the data distributions with ground-truth labels,
while plots~(b) and (c) are the classification results of Vanilla
LR and MFL-P-IU3, respectively. The red/blue stars and round points
represent the class(0/1) train and test data, while numbers in plots
are the classification accuracies of corresponding methods.} \label{fig:The-T-SNE-visulisation}
\end{figure}

\noindent \textbf{Visualisation}
To validate our hypothesis, i.e. the regularisation knowledge
transfer with MFL, we adopted T-SNE~\cite{maaten2008visualizing}
to visualise the classification results of Vanilla LR and MFL-P-IU3
on 2-way 1-shot tasks from the novel classes of \emph{mini}ImageNet.
Specifically, we showed three typical data distributions, i.e. continuity,
cluster and manifold, for comprehensively describing the regularisation
behaviors with the learned functional regularisation knowledge. Figure~\ref{fig:The-T-SNE-visulisation}
shows: (1) In a specific feature space, the data distributions fit
the characters of continuity, cluster or manifold (Fig.~\ref{fig:The-T-SNE-visulisation}(a));
(2) The few-shot classifiers easily overfit to the labelled data,
resulting in hypotheses lacking of regularisation and inferior classification
results (Fig.~\ref{fig:The-T-SNE-visulisation}(b)); (3) Our MFL-P-IU3
can remit this limitation via imposing the functional regularisation
knowledge into classifiers, achieving more reasonable hypotheses with
superior classification results (Fig.~\ref{fig:The-T-SNE-visulisation}(c)).

\noindent \textbf{Statistics of the improvements on novel classes. }
Essentially, our MFL methods aim to learn task-agnostic knowledge, i.e. the transferable and generalisable functional regularisation knowledge, to improve FSL classifiers.
Due to the functional regularisation knowledge is learned from episodes sampled from a base dataset, as a common learning favour of machine learning methods, the learned functional regularisation knowledge prefers to improve the FSL tasks containing the novel classes which are similar to the categories in a base dataset.  
To investigate this, 
% To investigate how MFL improves the performance on classifiers, we assume that this regularisation knowledge is closely related to the categories. 
%
% Therefore, 
we designed an experiment on binary classifiers whose parameters are closely related the positive class, and the learned functional regularisation may have different favours in improving binary classifiers with different positive classes.
Specifically, we employed the trained ComMFL-IU3 model to 2-way 1-shot FSL tasks from the novel classes of \emph{mini}ImageNet.
For each novel class, we randomly sampled 600 episodes containing one positive sample and one negative sample from other novel classes.

\begin{figure*}[t]
\centering
\includegraphics[scale=0.42]{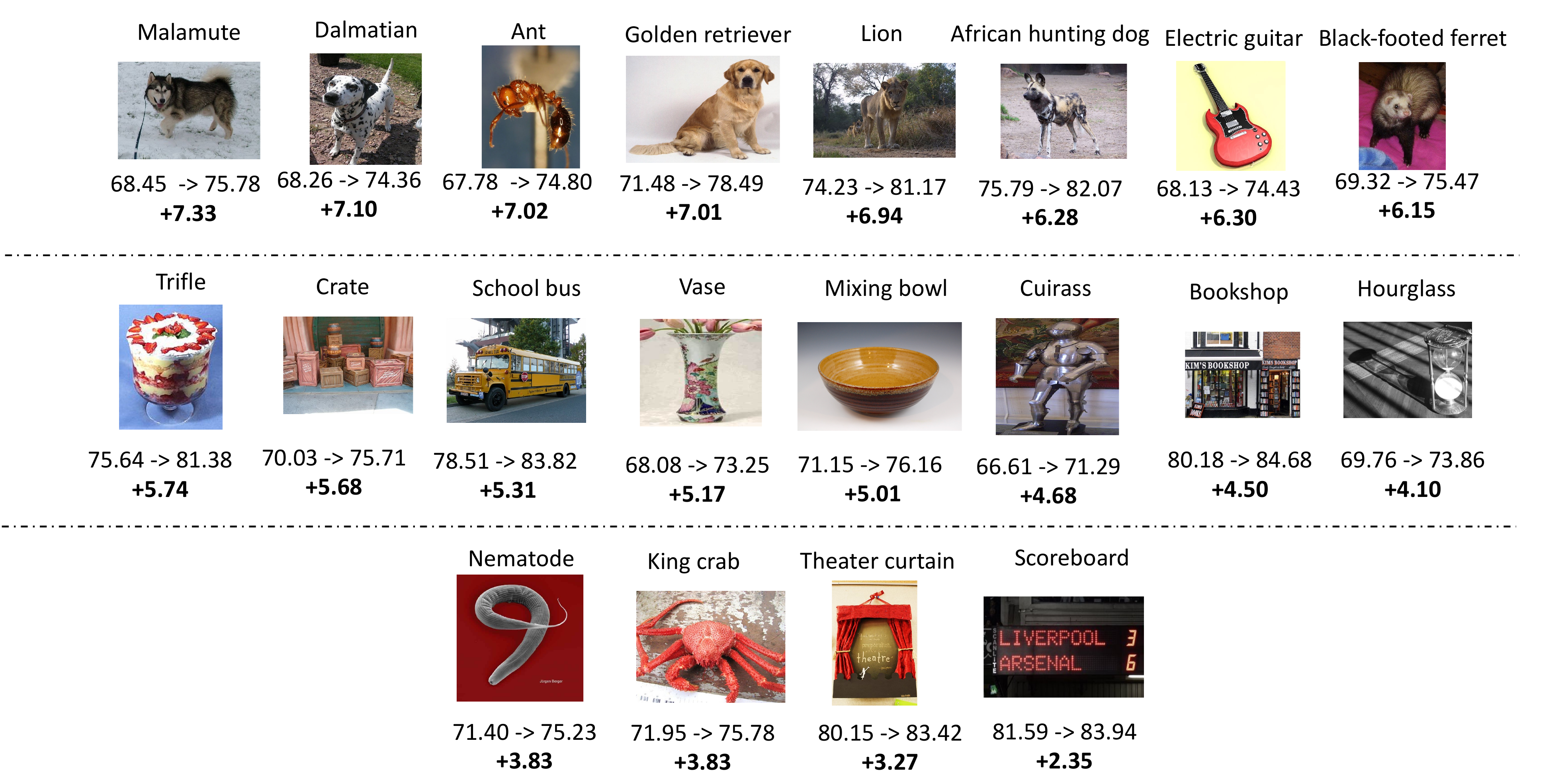}
\vspace{-0.1in}
\caption{Statistics of the improvements of ComMFL-IU3 on binary classification with various positive classes. The description of $a->b, + c$ represent the evaluation results on corresponding novel class, where $a$ and $b$ are the results using Vanilla LR and ComMFL-IU3, and $c$ is the improvement of ComMFL-IU3 compared with Vanilla LR.} \label{fig:statistics_improvement}
\end{figure*}

The statistics of the improvements on various novel classes are shown in Fig.~\ref{fig:statistics_improvement}.
We note that the learned functional regularisation performs well on the novel classes related to animal, i.e. Malamute, Dalmatian, Ant and Lion.
Moreover, the classes belonging to the dog category, i.e. Malamute, Dalmatian, Golden retriever and African hunting dog, show larger improvements compared with other classes.
This improvement may attribute to the related classes occurring in a base dataset, i.e. animals or other dog classes.
%
% In the contrast, the functional regularisation performs not as well as 
In general, however, all novel classes benefit from the learned functional regularisation, showing that this type of meta-knowledge is indeed useful for improving the FSL classifiers.

\noindent \textbf{Generalisation to different shots. }
\pan{
To demonstrate that our MFL methods are able to generalised to the FSL tasks with different shot, we conducted experiments on 5-way $K$-shot ($K={2,3,4,5}$) FSL with vanilla MFL, vanilla MFL-IU3, MFL-P-IU3 and ComMFL-IU3.
Table~\ref{tab:Accuracies-shots} shows the evaluation results
and we can see that:
(1) All the forms of MFL can boost the classifiers' performance on 2/3/4/5-shot FSL; 
(2) With the number of shot increasing, the improvement of our MFL methods over
Vanilla LR deceases.
This suggests that the hypotheses can gradually learn regularisation knowledge with the help of available
labelled data, yielding more robust hypotheses where the boosting space with regularisation knowledge is narrow, thus the learned functional regularisation knowledge brings less improvement.
}

\begin{table}[t]
\caption{Evaluation on 5-way $k$-shot from \emph{mini}ImageNet.
\textbf{Bold:} the best scores.}
\label{tab:Accuracies-shots}
\centering
\scalebox{0.9}
{
\begin{tabular}{lcccc}
\toprule
\#shot & 2-shot & 3-shot & 4-shot & 5-shot  \\
\midrule
Vanilla LR & 53.40$\pm$0.75 & 58.17$\pm$0.74 &60.93$\pm$0.73 & 62.98$\pm$0.71  \\
vanilla MFL& 56.09$\pm$0.76  &60.42$\pm$0.74 &62.86$\pm$0.73 &64.74$\pm$0.70   \\
    %  MFL-P & $\pm$0.88  & $\pm$0.87 & $\pm$0.75 & $\pm$0.75 &64.75$\pm$0.72 \\
vanilla MFL-IU3 & 58.01$\pm$0.78 & 62.27$\pm$0.74 &64.50$\pm$0.75 & 66.21$\pm$0.72  \\
     MFL-P-IU3  & 58.14$\pm$0.77 & 62.15$\pm$0.75 & 64.35$\pm$0.73 & 66.47$\pm$0.72    \\
ComMFL-IU3 & \textbf{58.84$\pm$0.78} & \textbf{63.26$\pm$0.75} &\textbf{65.19$\pm$0.74} & \textbf{67.18$\pm$0.72}    \\
\bottomrule
\end{tabular}
}
% \vspace{-0.35cm}
\end{table}

\noindent \textbf{Generalisation to different backbones}
\pan{We conducted experiments to investigate the generalisation
ability of our MFL methods on different backbones.
Specifically, we additionally used two backbone networks, i.e. ResNet12 in~\cite{wang2020instance} and recently proposed Shifted \emph{win}dow Transformer (Swin Transformer)~\cite{liu2021swin} for learning a representator.
In particular, we adopt the small version of Swin Transformer (Swin-S) with the default hyper-parameters in~\cite{liu2021swin} and the image size is resized as 224 $\times$ 224.
As in Tab.~\ref{tab:Mean-Accuracies-of-backbones}, our methods show well generalisation ability on different backbones.
Noticeable, the improvement on ResNet12 and Swin Transformer is smaller than that on Conv4, we conjecture this may attribute to the shallow architecture of
Conv4, yielding less discriminative representation in which the learned vanilla classifiers are easily stuck in the
overfitting problem and our MFL can effectively extricate them from this dilemma via the knowledge of functional regularisation.
}

\begin{table}[t]
\caption{Mean accuracies (\%) of Vanilla LR and LR with MFL and MFL-IU3 on
5-way 1/5shot tasks from \emph{mini}ImageNet.
\textbf{Bold:} the best scores.}
\label{tab:Mean-Accuracies-of-backbones}
\centering
\scalebox{0.96}
{
\begin{tabular}{lcccc}
\toprule 
{\small{}{}Backbone} & \multicolumn{2}{c}{{\small{}{}ResNet-12}} & \multicolumn{2}{c}{{\small{}{}Swin-Transfomer}}\tabularnewline
% \cmidrule{2-5}
{\small{}{}\#shot} & {\small{}{}1} & {\small{}5} & {\small{}{}1} & {\small{}{}5}\tabularnewline
\midrule 
{\small{}{}Vanilla LR} & {\small{}58.05} & {\small{}77.07} & {\small{}58.56} & {\small{}75.37}\tabularnewline
{\small{}{}vanilla MFL (Ours)} & {\small{}59.45} & {\small{}77.49} & {\small{}59.15} & {\small{}\textbf{75.52}}\tabularnewline
{\small{}{}vanilla MFL-IU3 (Ours)} & {\small{}60.40} & {\small{}77.24} & {\small{}59.35} & {\small{}75.35}\tabularnewline
% {\small{}{}MFL-P (Ours)} & {\small{}59.73} & {\small{}} & {\small{}59.18} & {\small{}--}\tabularnewline
{\small{}{}MFL-P-IU3 (Ours)} & \textbf{\small{}60.46} & \textbf{\small{}77.81} & \textbf{\small{}59.43} & {\small{}75.34}\tabularnewline
{\small{}{}ComMFL-IU3 (Ours)} & {\small{}60.24} & {\small{}77.20} & {\small{}59.14} & \textbf{\small{}75.50}\tabularnewline
\bottomrule
\end{tabular}
}
\end{table}

\noindent \textbf{Effects on different classifiers. }
\pan{
Essentially, the functional regularisation knowledge improves the FSL classifiers by imposing transferable constraints, and this type of knowledge should be, in principle, generalised to other parametric-classifiers and not limited to the Logistic Regression.
With this motivation, we conducted experiments to investigate the generalisation ability of MFL on different base classifiers.
Specifically, we additionally used linear Support Vector Machine (SVM) as a base classifier for learning a representation learned with Conv4 and ResNet12.
As expected, the results in Tab.~\ref{tab:Mean-Accuracies-of-classifiers} indicates that all forms of MFL show clear and consistent improvements over the Vanilla SVM, verifying the generalisation ability of our methods on different classifiers.
}

\begin{table}[t]
\caption{Mean accuracies (\%) of Vanilla SVM and SVM with MFL and MFL-IU3 on
5-way 1/5shot tasks from \emph{mini}ImageNet.
\textbf{Bold:} the best scores.}
\label{tab:Mean-Accuracies-of-classifiers}
\centering{}
% \centering {\small{}{}}%
\begin{tabular}{lcccc}
\toprule 
{\small{}{}Backbone} & \multicolumn{2}{c}{{\small{}{}Conv4}} & \multicolumn{2}{c}{{\small{}{}ResNet-12}}\tabularnewline 
% \cmidrule{2-5}
{\small{}{}\#shot} & {\small{}{}1} & {\small{}5} & {\small{}{}1} & {\small{}{}5}\tabularnewline
\midrule 
{\small{}{}Vanilla SVM} & {\small{}46.00} & {\small{}62.36} & {\small{}57.88} & {\small{}75.17}\tabularnewline
{\small{}{}vanilla MFL (Ours)} & {\small{}48.85} & {\small{}64.61} & {\small{}57.96} & {\small{}75.41}\tabularnewline
% {\small{}{}MFL-P (Ours)} & {\small{}49.00} & {\small{}64.39} & {\small{}58.95} & {\small{}74.81}\tabularnewline
{\small{}{}vanilla MFL-IU3 (Ours)} & {\small{}48.85} & {\small{}64.61} & {\small{}58.95} & {\small{}75.32}\tabularnewline
{\small{}{}MFL-P-IU3 (Ours)} & {\small{}51.25} & {\small{}65.56} & \textbf{\small{}60.59} & \textbf{\small{}76.26}\tabularnewline
{\small{}{}ComMFL-IU3 (Ours)} & \textbf{\small{}51.87} & \textbf{\small{}66.80} & {\small{}59.97} & {\small{}75.71}\tabularnewline
\bottomrule
\end{tabular}
\end{table}

\noindent \pan{\textbf{Influence of iterative steps. }
As the extensive experimental results in Tab.~\ref{tab:Mean-Accuracies-of-maintable} and Tab.~\ref{tab:Mean-Accuracies-of-CD} show that vanilla MFL-IU3 performs superior to vanilla MFL on FSL and CD-FSL with the help of iterative updates strategy.
But how do the MFLs, i.e. vanilla MFL, MFL-P and ComMFL, perform when we employ more iterative updates.
To answer this question, we evaluated vanilla MFL, MFL-P and ComMFL with different iterative updates $x=\{0,1,2,3,4,5,6\}$ on two typical scenarios, i.e. 5-way 1-shot FSL tasks and 5-way 1-shot CD-FSL tasks.}

\begin{figure*}[htbp]
\centering
\subfigure[Results on \emph{mini}ImageNet with Conv4 backbone.] {
 \label{fig:iu-mini} 
\begin{minipage}[t]{0.4\linewidth}
\centering
\includegraphics[width=3in]{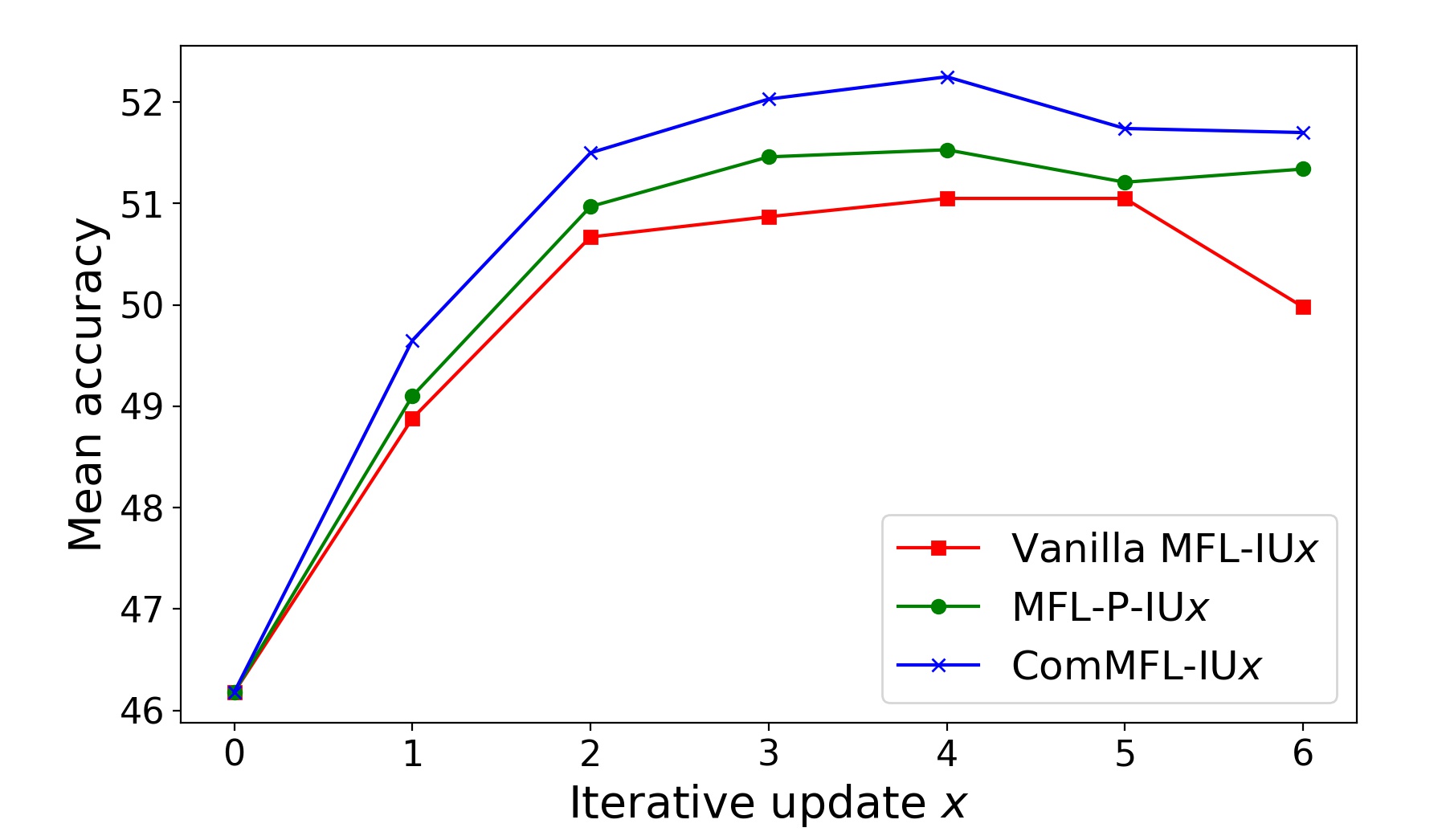}
\end{minipage}%
}%
\subfigure[Results under \emph{mini}ImageNet$\rightarrow$CUB scenario with Conv4 backbone] { 
\label{fig:iu-cub}
\begin{minipage}[t]{0.6\linewidth}
\centering
\includegraphics[width=3in]{./Figs/mini_iu_0to6.jpg}
%\caption{fig2}
\end{minipage}%
}%
\vspace{-0.05in}
\caption{Evaluation on 5-way 1-shot FSL tasks using the MFLs with $x$ iterative update. $x$=0 represents the result of Vanilla LR.}     
\label{fig:MFL_IUx}     
\end{figure*}

\pan{
Figure~\ref{fig:MFL_IUx} shows that, as expected, the performance gradually increases with the iterative updates $x$ increasing from 1 to 3.
However, when $x$ is too large, i.e. $x>4$ for FSL and $x>3$ for CD-FSL, the performance become stable even decreased.
%
% This founding inspires us to use $3$ as the default value for vanilla MFL-IU$x$, MFL-P-IU$x$ and ComMFL-IU$3$.
%
Noticeably, the best number of updates for CD-FSL is $3$, which is small than that for FSL.
We conjecture that this is due to the transferable regularisation knowledge across domain is less than that within domain, requiring less model capacity provided by the connected blocks with iterative updates.
}

\section{Conclusion}
In this work, we explored the idea of knowledge transfer by learning a meta functional of regularisation in the model learning functional spaces between a richly labelled domain and a scarcely labelled domain.
We demonstrate that classifiers with less training data can gradually learn the functional regularisation knowledge from a concurrent learning process on more labelled data. 
Based on this observation, we consider that this functional regularisation knowledge can be transferred across
different domains for model learning tasks when training data is scarce.
We formulated the MFL framework and generalised it to three different forms, i.e. a MFL with prototypes (MFL-P), a Composite MFL (ComMFL) and a MFL with Iterative Updates (MFL-IU). Extensive experiments on \emph{mini}ImageNet,
\emph{CIFAR-FS}, CUB, Cars and Places, show that the transfer of model learning
regularisation knowledge is effective in learning more accurate hypotheses
(classifiers) given scarcely labelled data.

% or


% % use section* for acknowledgment
% \ifCLASSOPTIONcompsoc
%   % The Computer Society usually uses the plural form
%   \section*{Acknowledgments}
% \else
%   % regular IEEE prefers the singular form
% \fi


% \ifCLASSOPTIONcaptionsoff
%   \newpage
% \fi

\bibliographystyle{ieee_fullname}
\bibliography{egbib}

\end{document}